%% file: main.tex
\documentclass[10pt,twocolumn,letterpaper]{article}

\usepackage[accsupp]{axessibility}
\usepackage[pagenumbers]{cvpr} 

\input{preamble}

\definecolor{cvprblue}{rgb}{0.21,0.49,0.74}
\usepackage[pagebackref,breaklinks,colorlinks,allcolors=cvprblue]{hyperref}

\newcolumntype{x}[1]{>{\centering\arraybackslash}p{#1pt}}
\newcolumntype{y}[1]{>{\raggedright\arraybackslash}p{#1pt}}
\newcolumntype{z}[1]{>{\raggedleft\arraybackslash}p{#1pt}}
\def\pz{{\phantom{0}}}
\newcommand{\grayrow}{\rowcolor[gray]{.95}}

\definecolor{baselinecolor}{gray}{.95}

\newcommand{\resolved}[3][]{\ifstrequal{#1}{resolved}{\textcolor{blue}{RESOLVED:}~\textbf{{\MakeUppercase #2:}}~{#3}}{\textbf{\MakeUppercase #2:}~#3}}

\newcommand{\wotext}{Baseline (w/o text)\xspace}

\newcommand\blfootnote[1]{\begingroup\renewcommand\thefootnote{}\footnote{#1}\addtocounter{footnote}{-1}\endgroup}

\title{Language-Guided Image Tokenization for Generation}

\author{
Kaiwen Zha\textsuperscript{1,2}\textsuperscript{*} \quad \quad 
Lijun Yu\textsuperscript{1}\quad \quad 
Alireza Fathi\textsuperscript{1}\quad \quad 
David A. Ross\textsuperscript{1}\\
Cordelia Schmid\textsuperscript{1}\quad \quad 
Dina Katabi\textsuperscript{2}\quad \quad 
Xiuye Gu\textsuperscript{1} \\
\textsuperscript{1}Google DeepMind \quad \quad
\textsuperscript{2}MIT CSAIL
}

\begin{document}
\input{figure_src/recon_samples}
\blfootnote{\textsuperscript{*} Work done during an internship at Google DeepMind.}

\vspace{-4mm}
\input{sec/0_abstract}
\input{sec/1_intro}
\input{sec/2_related_work}
\input{sec/3_method}
\input{sec/4_experiment}
\input{sec/5_conclusion}
\input{sec/6_acknowledgement}
{
    \small
    \bibliographystyle{ieeenat_fullname}
    \bibliography{main}
}
\input{sec/X_suppl}

\end{document}

%% file: preamble.tex
\usepackage[table]{colortbl}
\usepackage{multirow}
\usepackage{float}
\usepackage{xcolor}

%% file: figure_src/recon_samples.tex
\twocolumn[{%
\renewcommand\twocolumn[1][]{#1}%
\maketitle
\vspace{-4.5mm}
\includegraphics[width=\linewidth]{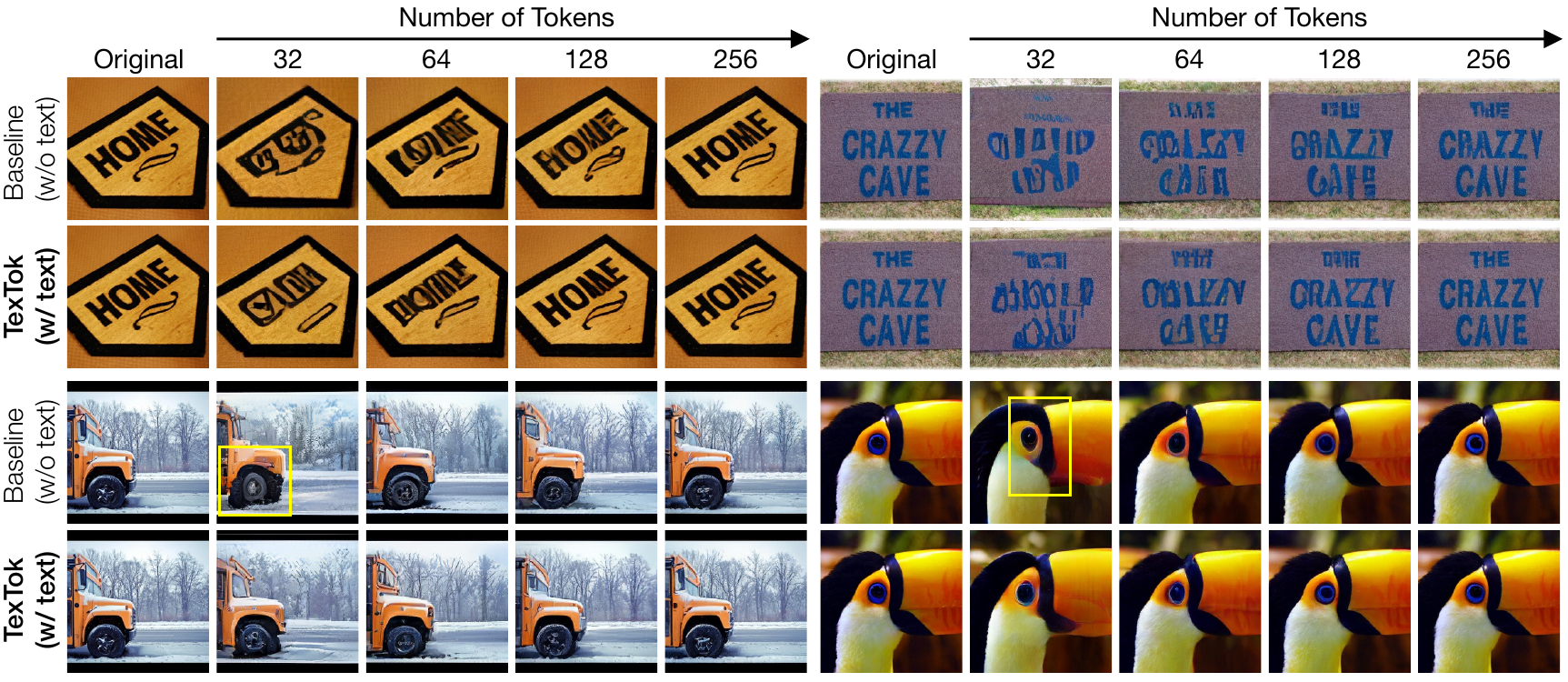}
\vspace{-7mm}
\captionof{figure}{\textbf{Reconstruction samples of TexTok compared with \wotext{}} on ImageNet 256$\times$256 using different number of image tokens. TexTok enables the tokenizer to encode finer visual details into image tokens, achieving better reconstruction quality across various token counts, such as improved text in images, car wheels, and bird beaks. The improvement is particularly significant in the low-token domain. The yellow-boxed regions highlight the significant enhancements.}
\label{fig:recon_samples}
\vspace{4mm}
}]

%% file: sec/0_abstract.tex
\begin{abstract}
\vspace{-6.5mm}

Image tokenization, the process of transforming raw image pixels into a compact low-dimensional latent representation, has proven crucial for scalable and efficient image generation. However, mainstream image tokenization methods generally have limited compression rates, making high-resolution image generation computationally expensive. To address this challenge, we propose to leverage language for efficient image tokenization, and we call our method Text-Conditioned Image Tokenization (TexTok). 
TexTok is a simple yet effective tokenization framework that leverages language to provide a compact, high-level semantic representation. By conditioning the tokenization process on descriptive text captions, TexTok simplifies semantic learning, allowing more learning capacity and token space to be allocated to capture fine-grained visual details, leading to enhanced reconstruction quality and higher compression rates.
Compared to the conventional tokenizer without text conditioning, TexTok achieves average reconstruction FID improvements of 29.2\% and 48.1\% on ImageNet-256 and -512 benchmarks respectively, across varying numbers of tokens. These tokenization improvements consistently translate to 16.3\% and 34.3\% average improvements in generation FID. 
By simply replacing the tokenizer in Diffusion Transformer (DiT) with TexTok, our system can achieve a
93.5$\times$ inference speedup while still outperforming the original DiT using only 32 tokens on ImageNet-512. TexTok with a vanilla DiT generator achieves state-of-the-art FID scores of 1.46 and 1.62 on ImageNet-256 and -512 respectively. 
Furthermore, we demonstrate TexTok's superiority on the text-to-image generation task, effectively utilizing the off-the-shelf text captions in tokenization. Project page is at: \renewcommand{\ttdefault}{pcr}{\small\url{https://kaiwenzha.github.io/textok/}}.
\end{abstract}

%% file: sec/1_intro.tex
\vspace{-4mm}
\section{Introduction}
\label{sec:intro}
Image generation has made remarkable progress in recent years, enabling high-quality synthesis across diverse applications~\cite{esser2021taming, rombach2022high, peebles2023scalable, dhariwal2021diffusion}. Central to this success is the evolution of image tokenization, a process that compresses raw image data into a compact yet expressive latent representation through training an autoencoder. Tokenization allows generative models, such as diffusion models~\cite{rombach2022high, peebles2023scalable, dhariwal2021diffusion} and autoregressive models~\cite{esser2021taming, tian2024visual, li2024autoregressive} to operate directly in this compressed latent space instead of the high-dimensional pixel space,  significantly improving computational efficiency while enhancing generation quality and fidelity.

Despite various image tokenization efforts aimed at improving training objectives~\cite{van2017neural,razavi2019generating,esser2021taming} and refining the autoencoder architecture~\cite{yu2021vector,yu2024an}, current methods remain fundamentally limited by a trade-off between compression rate and reconstruction quality, especially in high-resolution generation. High compression reduces computational costs but often sacrifices reconstruction quality, while prioritizing quality leads to prohibitively high computational expenses.

Addressing this limitation requires a fundamental shift in the tokenization process. At its core, tokenization involves finding a compact and effective representation of an image. The most concise and meaningful representation of an image often comes from its language description--i.e., captioning. When describing an image, humans naturally start with high-level semantics before elaborating on finer details. Inspired by this insight, we introduce \textit{Text-Conditioned Image Tokenization (TexTok)}, a novel framework that leverages text captions to guide the tokenizer in learning image semantics. This simplifies semantic learning, allowing more learning capacity and token space to be allocated to capture fine-grained visual details, thereby enhancing reconstruction quality without compromising compression rate.

To the best of our knowledge, we are the \textbf{first} to condition on detailed captions in the tokenization stage, an approach typically reserved for the generation phase. 
Text captions are easy to obtain from online image-text pairs or using a vision-language model to caption the images. Since text conditioning is widely used in image generation, e.g., text-to-image generation, our method can seamlessly incorporate these captions into the tokenization process without incurring additional annotation overhead.

We demonstrate the effectiveness of TexTok across a diverse set of tasks and settings. Compared to conventional tokenizers without text conditioning, TexTok achieves substantial gains in reconstruction quality, with average reconstruction FID improvements of \textbf{29.2\%} and \textbf{48.1\%} on ImageNet 256$\times$256 and 512$\times$512 resolutions, respectively. These enhancements in tokenization lead to consistent boosts in generation performance, with average improvements of \textbf{16.3\%} and \textbf{34.3\%} in generation FID for the two resolutions. By simply replacing the tokenizer in Diffusion Transformer (DiT) with TexTok, our system achieves a \textbf{93.5$\times$} inference speedup while still outperforming the original DiT using only \textbf{32} tokens on ImageNet 512$\times$512. Our best TexTok variant with a vanilla DiT generator achieves state-of-the-art FID scores of \textbf{1.46} and \textbf{1.62} on ImageNet 256$\times$256 and 512$\times$512 respectively.

We further demonstrate that incorporating text during the tokenization stage significantly enhances \textbf{text-to-image} generation, achieving 2.82 FID and 29.23 CLIP score on ImageNet 256$\times$256. Since text captions are inherently available for this task, TexTok boosts performance without adding any extra annotation overhead.

%% file: sec/2_related_work.tex
\section{Related Work}
\label{sec:related_work}
\paragraph{Image tokenization.} 
Image tokenizers build a bidirectional mapping between high-resolution pixels and a low-dimensional latent space, significantly improving the learning efficiency of downstream tasks, such as image generation~\cite{esser2021taming,chang2022maskgit,lee2022autoregressive,yu2024language}, and understanding~\cite{yu2021vector,yu2024language}.
Image tokenizers are usually formulated as an AutoEncoder (AE)~\cite{ballard1987modular} framework with an optional quantizer~\cite{van2017neural} and potentially in a variational~\cite{kingma2013auto} setup.
These AutoEncoders are trained to minimize the discrepancy between the output and input images, measured by pixel-space distances, latent-space distances~\cite{zhang2018unreasonable}, or jointly trained discriminators~\cite{esser2021taming}.
Architectural variants for the encoder and decoder include ResNet~\cite{he2016deep} and vision transformers~\cite{dosovitskiy2021an}.
Spatial correspondence has been a common property of modern tokenizer designs, where one token largely refers to a square neighborhood of pixels.
Recently, there has also been development of transformer-based models producing global tokens as a more compact representation~\cite{yu2024an}. In this work, we follow this paradigm to tokenize an image into a set of global tokens to flexibly control token budgets. However, unlike prior work, we are the first to propose to condition the tokenization process on image captions, which greatly improves the reconstruction quality and compression rate.
\vspace{-4mm}
\paragraph{Image generation.}
Generative learning of pixels has been explored under adversarial~\cite{brock2018large,sauer2022stylegan}, autoregressive~\cite{chen2020generative}, and diffusion~\cite{dhariwal2021diffusion,hoogeboom2023simple,kingma2024understanding} setups.
For higher resolutions, generative learning in compressed latent spaces has become popular given its efficiency advantages. Among them, autoregressive~\cite{esser2021taming,lee2022autoregressive} and masked prediction~\cite{chang2022maskgit,yu2024language} models often operate in discrete token spaces following the practice of GPT~\cite{radford2018improving} and BERT~\cite{Devlin2019BERTPO} in language modeling.
Recent variants~\cite{li2024autoregressive} could also use continuous latent spaces, akin to those used in latent diffusion models (LDMs)~\cite{rombach2022high}.
For LDMs, the architecture has evolved from convolution-based U-Net~\cite{ronneberger2015u} to transformer-based DiT~\cite{peebles2023scalable}.
In this paper, we focus on diffusion-based image generation with DiT architecture, leveraging the flexible token lengths of TexTok.
\vspace{-4mm}
\paragraph{Leveraging external semantic information in image generation and tokenization.}
Many recent studies start to leverage external semantic information, such as image representations and semantic maps, to improve image generation~\cite{pernias2023wurstchen, li2023self, yu2024representation}. Unlike these methods, which use external semantics to aid the generation process, our approach focuses on enhancing the tokenization process through conditioning on text semantics. Some recent efforts~\cite{yu2024spae, liu2024language, zhu2024beyond, liang2024lg} also consider aligning image tokens with text semantics in image tokenization to improve multimodal understanding. They either directly map images to text tokens in a frozen LLM codebook~\cite{yu2024spae, liu2024language, zhu2024beyond} or align the features of image tokens with text features~\cite{liang2024lg}, to produce semantically meaningful tokens. However, by enforcing strict image-text alignments, these works suffer from limited image reconstruction quality due to the inherent divergence between vision and language representations, resulting in undesirable image generation quality. In contrast, our work takes a complementary approach. We leverage text as external semantic conditioning, significantly boosting the image reconstruction and generation performance.

%% file: sec/3_method.tex
\section{Method}
\subsection{Preliminary}
Based on the format of latent representation, image tokenizers can be broadly classified into: 1) \textit{Vector-Quantized (VQ) Tokenizers}, such as VQ-VAE~\cite{van2017neural} and VQGAN~\cite{esser2021taming}, which represent images using a set of discrete tokens, and 2) \textit{Continuous Latent Tokenizers}~\cite{rombach2022high} which use a variational autoencoder (VAE)~\cite{kingma2013auto} to embed images into a continuous latent space. In this work, we focus primarily on continuous latent tokenizers. As shown in Appendix~\ref{app:discrete}, TexTok also works well on VQ tokenizers.

The standard continuous latent tokenizer typically consists of an encoder (tokenizer) $E$ and a decoder (detokenizer) $D$. Given an image $\mathbf{I}\in \mathbb{R}^{H\times W\times 3}$, the encoder $E$ compresses it into a 2D latent space $\mathbf{Z} = E(\mathbf{I}) \in \mathbb{R}^{h\times w\times d}$, where $h=\frac{H}{f}$, $w=\frac{W}{f}$, and $f$ is the spatial downsampling factor. Each latent embedding $\mathbf{z}\in \mathbb{R}^d$ is treated as a continuous token, with the image represented by a total of $hw$ tokens. For decoding, these embeddings $\mathbf{Z}$ are fed into the decoder $D$ to reconstruct the image $\hat{\mathbf{I}} = D(\mathbf{Z})$. Recently, 1D tokenizers~\cite{yu2024an} were introduced to allow  flexible token budgets for image representation, directly compressing $\mathbf{I}$ into 1D latent embeddings $\mathbf{Z}_{1D} = E(\mathbf{I}) \in \mathbb{R}^{N\times d}$ with $N$ tokens. Reconstruction, perceptual~\cite{zhang2018unreasonable}, and GAN~\cite{esser2021taming} losses are applied to train the tokenizer by minimizing the distance between $\mathbf{I}$ and $\hat{\mathbf{I}}$.

In this work, we adopt the 1D tokenizer paradigm to allow more flexible compression rates, demonstrating TexTok's efficacy and efficiency across varying token budgets. 

\input{figure_src/framework}
\subsection{TexTok: Text-Conditioned Image Tokenization}
We introduce \textit{Text-Conditioned Image Tokenization (TexTok)}, a simple yet effective tokenization framework. Unlike existing methods that compress all visual information into latent tokens, we use text captions to represent high-level semantics and guide the tokenization process. 

\vspace{-4.5mm}
\paragraph{Tokenization stage.} Given an image caption, we use a frozen T5~\cite{raffel2020exploring} text encoder to extract text embeddings. These embeddings are injected into both the tokenizer and detokenizer to providing semantic guidance throughout the tokenization process.

As shown in Figure~\ref{fig:framework}, TexTok adopts a Vision Transformer (ViT) backbone for both the encoder (tokenizer) $E$ and the decoder (detokenizer) $D$ to enable flexible control of token numbers. The input to the tokenizer is a concatenation of three components: 1) image patch tokens $\mathbf{P}\in \mathbb{R}^{hw\times D}$ from patchifying and flattening the input image with a projection layer, where $h=\frac{H}{s}$, $w=\frac{W}{s}$, and $s$ is the patch size, 2) $N$ randomly-initialized learnable image token $\mathbf{L} \in \mathbb{R}^{N\times D}$, where $N$ is the number of output image tokens, and 3) \textit{linearly projected text tokens}, $\mathbf{T}\in \mathbb{R}^{N_t\times D}$, derived from the text embeddings, where $N_t$ is the number of text tokens. In the tokenizer's output, only the learned image tokens are retained and linearly projected to produce the output image tokens $\mathbf{Z}\in\mathbb{R}^{N\times d}$.

The detokenizer also takes three concatenated inputs: 1) $hw$ learnable patch tokens $\mathbf{P'}\in \mathbb{R}^{hw\times D}$, 2) linearly projected image tokens $\mathbf{Z'}\in \mathbb{R}^{N\times D}$ from the input image tokens, and 3) \textit{linearly projected text tokens} $\mathbf{T'}\in \mathbb{R}^{N_t\times D}$ that come from the same text tokens fed to the tokenizer. In the detokenizer’s output, only the learned image patch tokens are retained, unpatchified, and projected to reconstruct the image patches.

We train the tokenizer and detokenizer using the combination of $\ell_2$ reconstruction, GAN, perceptual, and LeCAM regularization~\cite{tseng2021regularizing} losses, following~\cite{yu2024language}.

By directly injecting text tokens containing high-level semantic information into both the tokenizer and detokenizer, TexTok alleviates the need for the tokenizer and detokenizer to learn image semantics.

\vspace{-4mm}
\input{tables/tab-imagenet-256-512-main}
\paragraph{Generation stage.} Since this work focuses on continuous latent tokens, we use the Diffusion Transformer (DiT)~\cite{peebles2023scalable} as the generation framework and train the DiT on top of the latent tokens produced by TexTok. 
Note that \textit{only latent image tokens} need to be generated in the generation stage, while the text tokens will be provided in detokenization. 

DiT is trained to model the distribution of TexTok latent tokens, conditioned either on a class category (for class-conditional generation) or on the text embeddings (for text-to-image generation).

During inference, the process differs by the generation task. 
For text-to-image generation, we use the provided captions for both generation and detokenization, feeding the text embeddings and generated latent image tokens into the detokenizer to produce the output image.
For class-conditional generation, DiT generates latent tokens based on the specified class; we then sample an unseen caption for that class from a pre-generated list, and inject it into the detokenizer along with the generated latent tokens to produce the final image. Notably, only the class category is used during generation, in line with standard practice. 

%% file: figure_src/framework.tex
\begin{figure}[t]\centering
\includegraphics[width=\linewidth]{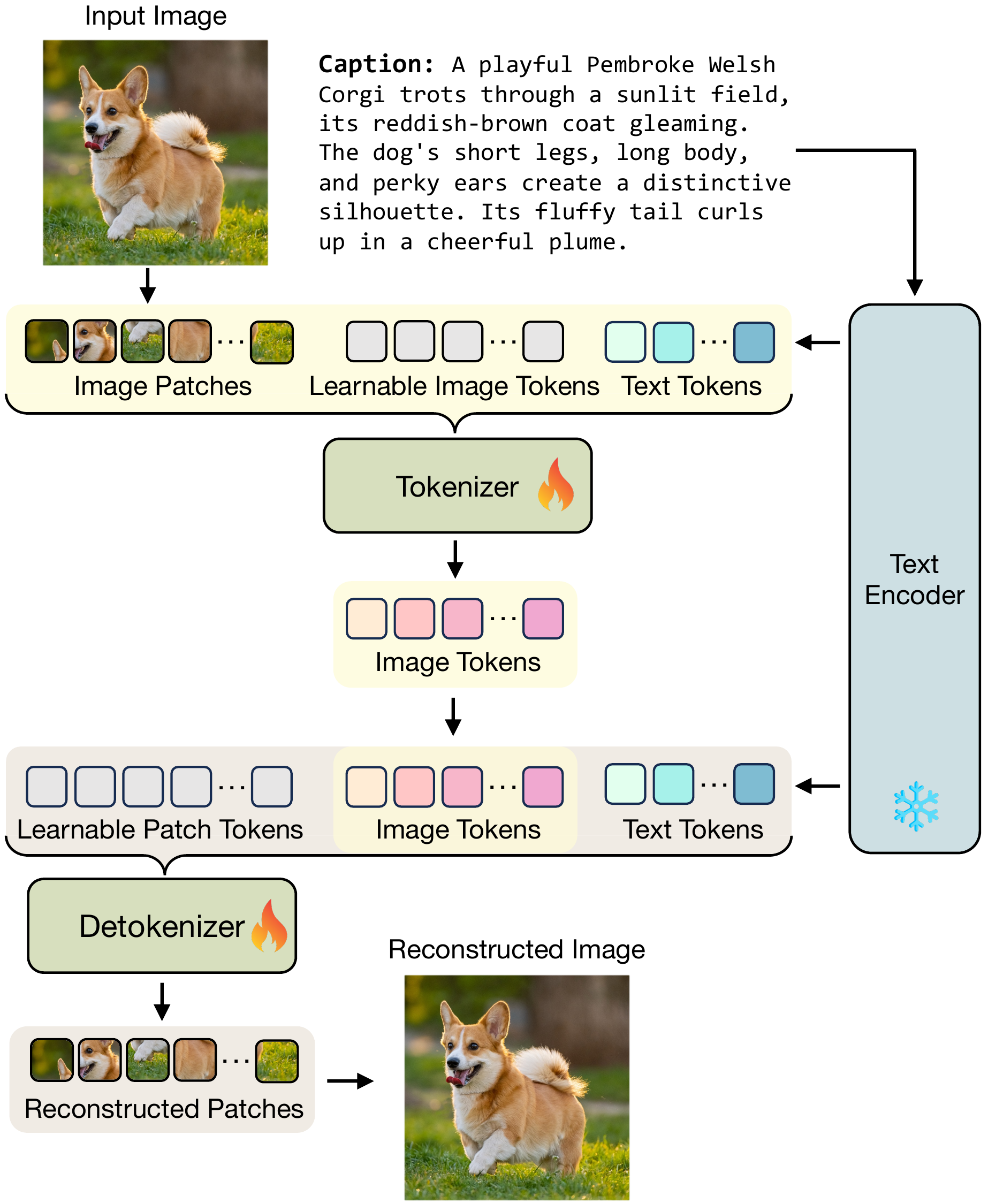}
\caption{\textbf{TexTok architecture}. {During training, a frozen text encoder (e.g., T5~\cite{raffel2020exploring}) extracts text embeddings (tokens) from the given image caption. The image patches, learnable image tokens, and text tokens are fed into the tokenizer (a ViT~\cite{dosovitskiy2021an}) to produce the image tokens. During detokenization, the image tokens are concatenated with the same text tokens fed to the tokenizer and learnable patch tokens to reconstruct the image. For generation, \textit{only image tokens} need to be generated.}}
\vspace{-4mm}
\label{fig:framework}
\end{figure}

%% file: tables/tab-imagenet-256-512-main.tex
\begin{figure*}[t]
\centering
\vspace{-3mm}
\begin{minipage}{0.6\textwidth}
\vspace{-3mm}
\begin{table}[H]
\footnotesize
    \begin{center}
    \scalebox{0.78}{
    \begin{tabular}{lcccccc|cc}
    \toprule
    \multicolumn{7}{c|}{\textbf{Reconstruction}} & \multicolumn{2}{c}{\textbf{Generation}} \\
    \midrule
    tokenizer & \# tokens & rFID $\downarrow$   & rIS$\uparrow$  & PSNR$\uparrow$     & SSIM $\uparrow$ & LPIPS$\downarrow$ & gFID $\downarrow$ & gIS $\uparrow$ \\
    \midrule\midrule
    \multicolumn{7}{l}{\textbf{(a) ImageNet 256$\times$256}} \\
    \arrayrulecolor{gray}\midrule
    \textcolor{gray}{SD-VAE-f8~\cite{rombach2022high}} & \textcolor{gray}{1024~\scriptsize{(d=4)}} & \textcolor{gray}{1.20$^\dagger$} & \textcolor{gray}{-} & \textcolor{gray}{-} & \textcolor{gray}{-} & \textcolor{gray}{-} & \textcolor{gray}{9.62} & \textcolor{gray}{121.5} \\
    \midrule
    Baseline-32~\scriptsize{(w/o text)} & \multirow{2}{*}{32~\scriptsize{(d=8)}} & 3.82 & 117.1 & 17.67 & 0.4281 & 0.3270 & 4.97 & 170.3 \\
    \textbf{TexTok-32~\scriptsize{(w/ text)}} & & \textbf{2.40} & \textbf{156.2} & \textbf{18.32} & \textbf{0.4463} & \textbf{0.2884} & \textbf{3.55} & \textbf{205.3} \\
    \midrule
    Baseline-64~\scriptsize{(w/o text)} & \multirow{2}{*}{64~\scriptsize{(d=8)}} & 2.04 & 147.2 & 19.52 & 0.4801 & 0.2343 & 3.30 & 188.9\\
    \textbf{TexTok-64~\scriptsize{(w/ text)}} & & \textbf{1.53} & \textbf{169.8} & \textbf{20.10} & \textbf{0.4971} & \textbf{0.2126} &\textbf{2.88} & \textbf{209.2} \\
    \midrule
    Baseline-128~\scriptsize{(w/o text)} & \multirow{2}{*}{128~\scriptsize{(d=8)}} & 1.49 & 160.5 & 20.51 & 0.5102 & 0.1913 & 3.19 & 190.1 \\
    \textbf{TexTok-128~\scriptsize{(w/ text)}} & & \textbf{1.04} & \textbf{183.3} & \textbf{22.05} & \textbf{0.5618} & \textbf{0.1499} & \textbf{2.75} & \textbf{210.9} \\
    \midrule
    Baseline-256~\scriptsize{(w/o text)} & \multirow{2}{*}{256~\scriptsize{(d=8)}} & 0.91 & 178.3 & 23.05 & 0.5950 & 0.1225 & 2.91 & 197.2 \\
    \textbf{TexTok-256~\scriptsize{(w/ text)}} & & \textbf{0.69} & \textbf{192.6} & \textbf{24.38} & \textbf{0.6454} & \textbf{0.0998} & \textbf{2.68} & \textbf{219.6} \\
    \midrule\midrule
    \multicolumn{7}{l}{\textbf{(b) ImageNet 512$\times$512}} \\
    \arrayrulecolor{gray}\midrule
    \textcolor{gray}{SD-VAE-f8~\cite{rombach2022high}} & \textcolor{gray}{4096~\scriptsize{(d=4)}} & \textcolor{gray}{-} & \textcolor{gray}{-} & \textcolor{gray}{-} & \textcolor{gray}{-} & \textcolor{gray}{-} & \textcolor{gray}{12.03} & \textcolor{gray}{105.3} \\
    \midrule
    Baseline-32~\scriptsize{(w/o text)} & \multirow{2}{*}{32~\scriptsize{(d=8)}} & 7.68 & 82.6 & 16.21 & 0.5046 & 0.4771 & 9.22 & 119.0\\
    \textbf{TexTok-32~\scriptsize{(w/ text)}} & & \textbf{2.33} & \textbf{161.5} & \textbf{18.55} & \textbf{0.5488} & \textbf{0.3772} & \textbf{3.61} & \textbf{215.6} \\
    \midrule
    Baseline-64~\scriptsize{(w/o text)} & \multirow{2}{*}{64~\scriptsize{(d=8)}} & 4.81 & 104.0 & 17.81 & 0.5341 & 0.4029 & 7.26 & 141.8\\
    \textbf{TexTok-64~\scriptsize{(w/ text)}} & & \textbf{1.52} & \textbf{171.7} & \textbf{20.19} & \textbf{0.5786} & \textbf{0.3093} & \textbf{3.30} & \textbf{210.2} \\
    \midrule
    Baseline-128~\scriptsize{(w/o text)} & \multirow{2}{*}{128~\scriptsize{(d=8)}} & 1.45 & 163.2 & 21.59 & 0.6086 & 0.2624 & 3.64 & 191.1 \\
    \textbf{TexTok-128~\scriptsize{(w/ text)}} & & \textbf{0.97} & \textbf{185.5} & \textbf{22.27} & \textbf{0.6230} & \textbf{0.2365} & \textbf{3.16} & \textbf{210.7} \\
    \midrule
    Baseline-256~\scriptsize{(w/o text)} & \multirow{2}{*}{256~\scriptsize{(d=8)}} & 1.07 & 174.9 & 23.15 & 0.6410 & 0.2180 & 3.14 & 204.2 \\
    \textbf{TexTok-256~\scriptsize{(w/ text)}} & & \textbf{0.73} & \textbf{192.0} & \textbf{24.45} & \textbf{0.6682} & \textbf{0.1875} & \textbf{2.87} & \textbf{218.5}\\
    \bottomrule
    \end{tabular}
    } %
    \end{center}
    \vspace{-5mm}
    \caption{\footnotesize \textbf{Image reconstruction and generation performance comparison of TexTok with \wotext{}} on ImageNet 256$\times$256 and 512$\times$512. 
    TexTok consistently delivers significant improvements in image reconstruction and generation performance, with more pronounced gains as the number of tokens decreases. Class-conditional generation results are reported without classifier-free guidance (Baseline and TexTok use DiT-L as the generator, while SD-VAE uses DiT-XL/2). $^\dagger$: number taken from~\cite{li2024autoregressive}.}  %
    \label{table:imagenet-256-512-main}
\end{table}
\end{minipage}%
\hfill
\begin{minipage}{0.38\textwidth}
    \vspace{2mm}
    \centering
    \subfloat[ImageNet 256$\times$256]{%
        \includegraphics[width=0.7\linewidth]{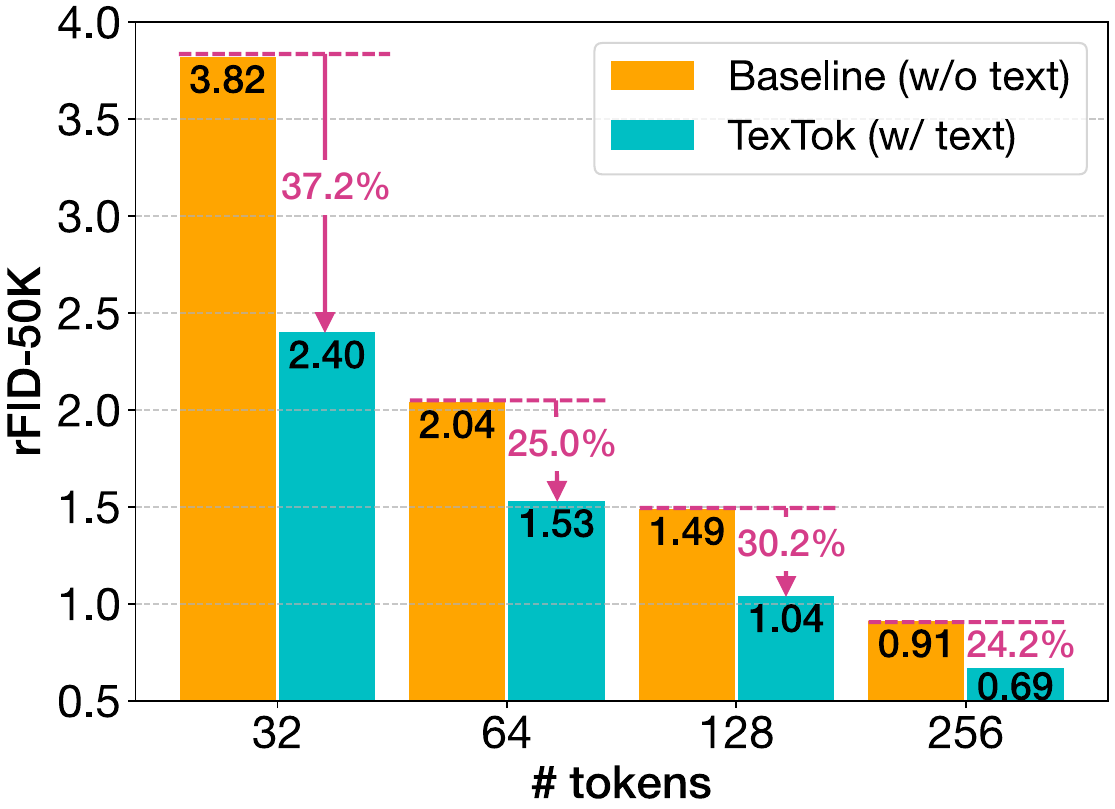}%
        \label{fig:w_wo_text_comparison_256}%
    }
    \\
    \vspace{1.6mm}
    \subfloat[ImageNet 512$\times$512]{%
        \includegraphics[width=0.7\linewidth]{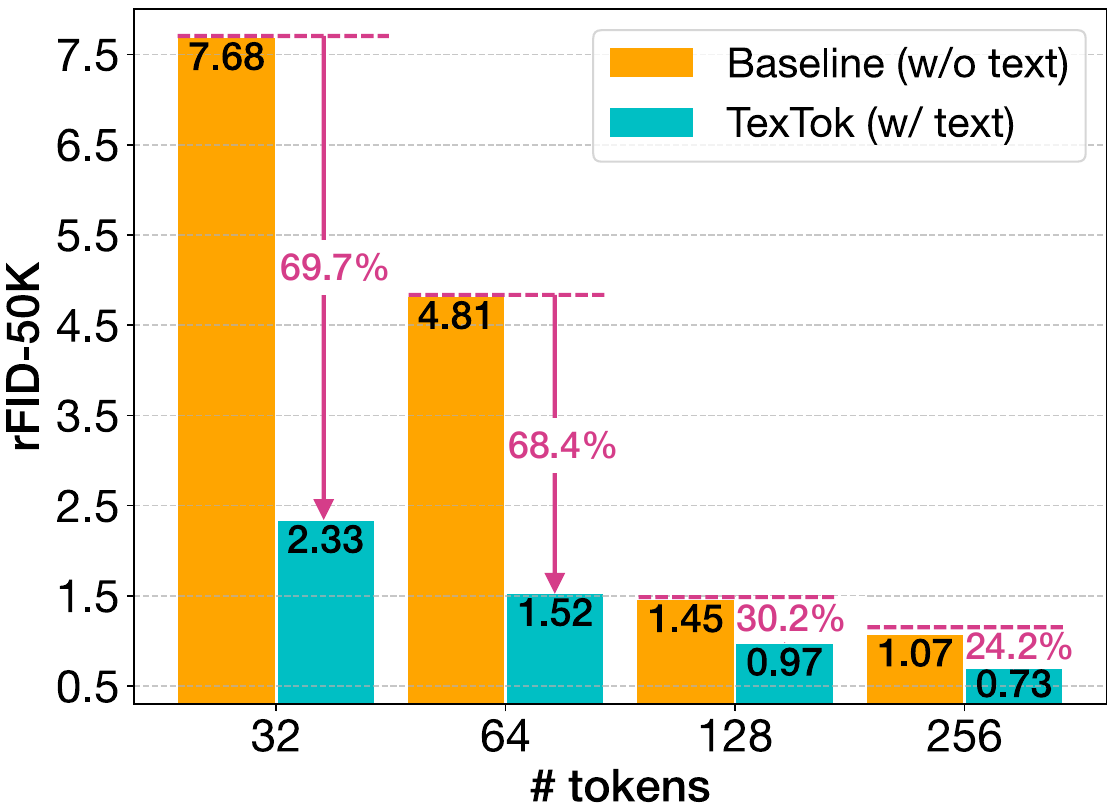}%
        \label{fig:w_wo_text_comparison_512}%
    }
    \vspace{-2mm}
    \caption{\footnotesize \textbf{Reconstruction FID of TexTok \textit{v.s.} \wotext{}} on ImageNet 256$\times$256 and 512$\times$512 for different number of image tokens. With text conditioning, TexTok can use \textbf{half}, \textbf{1/4} of the token number (\textbf{2$\times$}, \textbf{4$\times$} compression rates) to achieve similar rFID compared to \wotext{} on ImageNet-256 and -512 respectively.}
    \label{fig:w_wo_text_comparison_combined}
\end{minipage}
\vspace{-4mm}
\end{figure*}

%% file: sec/4_experiment.tex
\section{Experiments}
\subsection{Implementation Details} \label{sec:implementation}
The implementation details of TexTok are described below. Please refer to Appendix~\ref{app:implementation} for further details.

\vspace{-4mm}
\paragraph{Text caption acquision.} 
Text captions are readily available for text-to-image generation tasks, where they can be directly used in the tokenization process. For other generation tasks without captions, such as our use of ImageNet~\cite{deng2009imagenet}, we employ a vision language model (VLM), Gemini v1.5 Flash~\cite{team2024gemini}, to generate detailed captions offline. For the training set, we caption each given image. For the evaluation set, in class-conditional generation, we pre-generate unseen captions for each category using a sampled caption list of this category from the training set as reference. By default, each image is captioned with up to 75 words, which are encoded into a 128-token sequence using the T5 text encoder~\cite{raffel2020exploring} (XL for ImageNet-256 and XXL for ImageNet-512 experiments). Please see Appendix~\ref{app:captioning} for more details.

\vspace{-4.5mm}
\paragraph{Tokenization \& generation.} By default, all TexTok experiments employ ViT-Base for both tokenizer and detokenizer, each comprising 12 layers with a hidden size of 768 and 12 attention heads ($\sim$176M parameters). For the GAN loss, we follow~\cite{yu2021vector} and use the StyleGAN discriminator~\cite{karras2019style}($\sim$24M parameters). Unless otherwise specified,  the image token channel dimension in TexTok is set to $d=8$. 

We use Diffusion Transformer (DiT)~\cite{peebles2023scalable} as our default generator due to its effectiveness and flexibility of handling 1D tokens. We use a DiT patch size of 1 for all TexTok generation experiments, and by default, we train DiT for 350 epochs. Specifically, for class-conditional generation, we use the original DiT architecture. For text-to-image generation, referring to~\cite{chen2023pixart}, we modify DiT architecture by adding an additional multi-head cross-attention layer following the multi-head self-attention layer in the DiT block to accept text embeddings. We refer to this architecture as~``DiT-T2I". 

\subsection{Experiment Setup}
\paragraph{Model variants.} We compare two setups to demonstrate the effectiveness of using text conditioning: \emph{TexTok} incorporates text tokens in both the tokenizer and detokenizer, corresponding to the architecture shown in Figure~\ref{fig:framework}. In contrast, \emph{Baseline (w/o text)} does not condition on text tokens in both the tokenizer and detokenizer.
For each image, we tokenize it into ``\#tokens" number of latent tokens and train the generator to generate these tokens. 

\vspace{-4mm}
\paragraph{Evaluation protocol.} To evaluate reconstruction performance of the tokenizer, we report reconstruction Frechet inception distance (rFID)~\cite{heusel2017gans}, reconstruction inception score (rIS)~\cite{salimans2016improved}, peak signal-to-noise ratio (PSNR), structural similarity index measure (SSIM), and learned perceptual image patch similarity (LPIPS)~\cite{zhang2018unreasonable} on 50K samples from ImageNet training set. To evaluate class-conditional generation performance, we report generation Frechet inception distance (gFID)~\cite{heusel2017gans}, generation inception score (gIS)~\cite{salimans2016improved}, precision and recall~\cite{kynkaanniemi2019improved} following the evaluation protocol and suite provided by ADM~\cite{dhariwal2021diffusion}. To evaluate text-to-image generation performance, we report FID and CLIP Score~\cite{hessel2021clipscore} on 50K samples from ImageNet validation set.

\subsection{Effectiveness of Text Conditioning}
We begin by evaluating the effectiveness of text conditioning in image tokenization and generation. We compare our method, TexTok, with a \wotext{} that uses the same settings but excludes text conditioning, on ImageNet at resolutions of 256$\times$256 and 512$\times$512. We experiment with varying numbers of tokens, presenting the quantitative results in Table~\ref{table:imagenet-256-512-main} and visualizing the relative improvement in rFID in Figure~\ref{fig:w_wo_text_comparison_combined}.

On ImageNet 256$\times$256, across all settings, TexTok significantly enhances both reconstruction and generation performance. Specifically, TexTok achieves 37.2\%, 25.0\%, 30.2\%, 24.2\% improvements in rFID using 32, 64, 128, and 256 tokens respectively, which consistently translates to 28.6\%, 12.7\%, 13.8\%, and 7.9\% improvements in gFID. Notably, the fewer tokens used, the higher the gains from text conditioning. 
As shown in Figure~\ref{fig:w_wo_text_comparison_256}, TexTok can achieve similar rFID using \textbf{half} the number of tokens compared to the baseline \textbf{(2$\times$ compression rate)}. We note that our \wotext{} is highly competitive. As shown in Table~\ref{table:imagenet-256-512-main}(a), with 8$\times$ fewer number of tokens, \wotext{} outperforms the widely used SD-VAE tokenizer~\cite{rombach2022high} in both reconstruction and generation.

On higher resolution images, \ie, ImageNet 512$\times$512, TexTok exhibits stronger efficacy. As shown in Table~\ref{table:imagenet-256-512-main} and Figure~\ref{fig:w_wo_text_comparison_512}, TexTok achieves more significant improvement in the reconstruction quality and enables higher compression rates under this high-resolution setting. Specficially, it achieves 69.7\%, 68.4\%, 30.2\%, and 24.2\% improvements in rFID and 60.8\%, 54.5\%, 13.2\% and 8.6\% improvements in gFID, across 32, 64, 128 and 256 tokens respectively. As shown in Figure~\ref{fig:w_wo_text_comparison_512}, TexTok achieves similar rFID to the baseline using only \textbf{1/4 of the token number (4$\times$ compression rate)}. 

Finally, the qualitative results in Figure~\ref{fig:recon_samples} across varying token counts show that TexTok significantly enhances reconstruction quality, particularly for text within images and specific visual details, such as car wheels and beaks. This indicates that TexTok encodes finer visual details using the same number of tokens.

\subsection{System-level Image Generation Comparison}
We experiment with image generation using TexTok as the tokenizer and adopt a vanilla DiT image generator~\cite{peebles2023scalable} (denoted by TexTok + DiT), to study how this system performs against other leading image generation systems. 
We evaluate on class-conditional ImageNet 256$\times$256 and 512$\times$512 settings with varying number of tokens (compression rates).

On ImageNet 256$\times$256 class-conditional image generation, as shown in Table~\ref{tab:imagenet-system}(a), our TexTok-256 + DiT-XL achieves an FID of \textbf{1.46}, surpassing previous state-of-the-art systems, even though using a simpler, vanilla DiT as the image generator. As we reduce the number of tokens and increase image compression rate, TexTok + DiT maintains generation performance. Notably, TexTok-64 + DiT-XL, where the diffusion transformer generates only 64 image tokens, outperforms the original DiT-XL/2, which uses 256 tokens after patchification in the diffusion transformer.

\input{tables/tab-imagenet-256-512-system}
\input{figure_src/inference_time_256_512}
\input{figure_src/t2i_samples}
\input{figure_src/example_512}

On higher resolution images, \ie, ImageNet 512$\times$512, as shown in Table~\ref{tab:imagenet-system}(b), TexTok-256 + DiT-XL also achieves state-of-the-art \textbf{1.62} gFID compared with previous methods, using only \textbf{256} image tokens.
On the most compressed side, TexTok-32 + DiT-XL only uses 32 tokens yet achieves better generation performance than the original DiT that uses 1024 tokens after patchification.

Our system not only achieves superior generation performance, but is also very efficient given its great compression rates.
We plot in Figure~\ref{fig:inference_time_256} the speed v.s. performance tradeoffs of TexTok + DiT-XL compared to the original DiT on ImageNet 256$\times$256. Simply replacing the tokenizer in DiT with TexTok can achieve a \textbf{14.3$\times$ speedup} with better FID, or \textbf{34.3\% FID improvement} with similar inference time. This verifies the effectiveness and efficiency of TexTok.
This improved speed/performance tradeoff is further reflected on ImageNet 512$\times$512 (Figure~\ref{fig:inference_time_512}), where we demonstrate that simply replacing the tokenizer in DiT with TexTok variants, it achieves a \textbf{93.5$\times$} speedup with better FID using \textbf{32} tokens, or \textbf{46.7\%} FID improvement with \textbf{3.7$\times$} less inference time using 256 tokens. This shows that as image resolution increases, providing the tokenization process with explicit text semantics yields greater improvements in generation performance and inference speedup.

Qualitative samples in Figure~\ref{fig:teaser} demonstrate that TexTok enables class-conditional generation of semantically rich images with fine-grained details. More qualitative samples can be found in Appendix~\ref{app:qualitative}.

\input{tables/tab-ablation}
\subsection{Text-to-Image Generation}
We now demonstrate TexTok's superiority on text-to-image generation. We use the same VLM-generated captions on ImageNet 256$\times$256 with our modified DiT-T2I architecture (detailed in Section~\ref{sec:implementation}).  During training, the tokenizer and generator share the same text embeddings extracted by the T5 text encoder. During inference, we generate images condition on captions from ImageNet validation set. We calculate FID between these generated images and the original ImageNet validation set. As shown in Table~\ref{tab:t2i}, compared with \wotext{}, TexTok consistently and significantly improves text-to-image generation, across varying numbers of image tokens. 
Since text captions are already available for text-to-image tasks and the tokenizer can directly use the same text embeddings used in the generator, TexTok's performance boost comes at no additional cost for captioning and text embedding extraction.

Qualitative samples in Figure~\ref{fig:t2i_samples} show that TexTok's generation is more realistic and follows the prompts better. More qualitative samples can be found in Appendix~\ref{app:qualitative}.

\subsection{Tokenization/Generation Inference Efficiency}
We have demonstrated that TexTok significantly enhances reconstruction, class-conditional generation, and text-to-image generation quality. In text-to-image tasks, our text conditioning incurs no additional cost for text embedding extraction, as text embeddings are also used as conditioning in generation. For other tasks, it introduces minimal computational overhead to generate text embeddings and use them during tokenization. As shown in Table~\ref{tab:inference_time}, this overhead is negligible ($\sim$0.01 s/img). More importantly, the resulting reduction in generation computational cost compensates for this small increase, as evidenced by the comparison of computational costs between SD-VAE, \wotext{} and TexTok in Table~\ref{tab:inference_time} and the speedup results in Figure~\ref{fig:inference_time}.

\subsection{Ablation Studies}
We ablate TexTok to analyze the contribution of our design choices. We use the following default settings: TexTok-128, Base model size, and T5-XL text encoder. Captions are 75 words long and applied to both the tokenizer and detokenizer using in-context conditioning.

\vspace{-4mm}
\paragraph{Amount of text conditioning.} In Table~\ref{tab:cond_types}, we ablate various types of class/text conditioning: (1) a learnable class embedding based on the class category, (2) text embeddings from a short text template with class names, (3) text embeddings from 25-word captions, and (4) (ours) text embeddings from 75-word captions. Our results show that more descriptive text conditioning improves performance.

\vspace{-4mm}
\paragraph{T5 text encoder size.} In Table~\ref{tab:t5_size}, we study the effect of the text encoder model size. We find that a larger encoder leads to better reconstruction quality. We use T5-XL as the default setting on ImageNet-256 for its efficiency.

\vspace{-4mm}
\paragraph{Conditioning architecture.} Another design choice is how we inject text into the tokenizer and detokenizer. In Table~\ref{tab:cond_design}, we find that in-context conditioning  (concatenating text embeddings with other input tokens and feeding them into the self-attention layers) outperforms adding an additional multi-head cross-attention layer in each ViT block. 

\input{tables/tab-inference-time}
\vspace{-4.5mm}
\paragraph{Conditioning location.} In Table~\ref{tab:cond_location}, we ablate the locations for text conditioning injection and find that applying it to both the tokenizer and detokenizer yields the best results.

\vspace{-4.5mm}
\paragraph{TexTok model size.} In Table~\ref{tab:tokenizer_size}, we investigate the influence of TexTok model size. We find using TexTok-Base performs much better than TexTok-Small, but increasing the model size further provides marginal improvements. Hence, we choose TexTok-Base as our default model size. 

%% file: tables/tab-imagenet-256-512-system.tex
\setlength{\tabcolsep}{4pt}
\begin{table*}[!h]
    \begin{center}
    \scalebox{0.7}{
    \begin{tabular}{lcc|ccccc|ccccc}
    \toprule
    \multicolumn{3}{l|}{} & \multicolumn{5}{c|}{\textbf{(a) ImageNet 256$\times$256}} & \multicolumn{5}{c}{\textbf{(b) ImageNet 512$\times$512}} \\
    \midrule\midrule
    Model & \#Params (G) & \#Params (T) & FID$\downarrow$   & IS$\uparrow$ & Precision$\uparrow$ & Recall$\uparrow$ & \#tokens & FID$\downarrow$   & IS$\uparrow$ & Precision$\uparrow$ & Recall$\uparrow$ & \#tokens \\
    \midrule
    \multicolumn{6}{l}{\textit{GAN}\vspace{0.02in}} \\
    \pz\pz StyleGAN-XL~\cite{sauer2022stylegan} & 168M & - & 2.30 & 265.1 & 0.78 & 0.53 & - & 2.41 & 267.8 & 0.77 & 0.52  & - \\
    \arrayrulecolor{gray}\cmidrule(lr){1-13}
    \multicolumn{2}{l}{\textit{pixel diffusion}\vspace{0.02in}} \\
    \pz\pz ADM-U~\cite{dhariwal2021diffusion} &731M & - & 3.94 & 215.8 & 0.83 &0.53 & - & 3.85 & 221.7 & 0.84 &0.53 & -\\
    \pz\pz simple diffusion~\cite{hoogeboom2023simple} & 2B & - & 2.44 & 256.3 & - & -  & - & 3.02 & 248.7 & -& - & -\\
    \pz\pz VDM++~\cite{kingma2024understanding} & 2B & - & 2.12 & 267.7 & - & -  & - & 2.65 & 278.1 & - & - & -\\
    \arrayrulecolor{gray}\cmidrule(lr){1-13}
    \multicolumn{6}{l}{\textit{masked image modeling}\vspace{0.02in}} \\
    \pz\pz MaskGIT~\cite{chang2022maskgit} & 227M & 66M & 6.18 & 182.1 & 0.80 & 0.51 & 256 & 7.32 & 156.0 & 0.78 & 0.50 & 1024\\
    \pz\pz RCG~\cite{li2023self} & 512M & 66M & 2.25 & 300.7 & - & -  & 256 & - & - & - & - & -\\
    \pz\pz TiTok-L-32~\cite{yu2024an} & 177M & 644M & 2.77 & - & - & -  & 32 & - & - & - & - & -\\
    \pz\pz TiTok-64 (B/L)~\cite{yu2024an} & 177M & 202M / 644M& 2.48 & - & - & -  & 64 & 2.74 & - & - & - & 64\\
    \pz\pz TiTok-128 (S/B)~\cite{yu2024an} & 287M / 177M& 72M / 202M & 1.97 & - & - & -  & 128  & 2.13 & - & - & - & 128\\
    \pz\pz MAGVIT-v2~\cite{yu2024language} & 307M & 116M & 1.78 & 319.4 & - & -  & 256 & 1.91 & \textbf{324.3} & - & - & 1024\\
    \pz\pz MaskBit~\cite{weber2024maskbit} & 305M & 54M & 1.52 & \textbf{328.6} & - & -  & 256 & - & - & - & - & -\\
    \arrayrulecolor{gray}\cmidrule(lr){1-13}
    \multicolumn{6}{l}{\textit{autoregressive}\vspace{0.02in}} \\
    \pz\pz VQGAN~\cite{esser2021taming} & 1.4B & 23M & 15.78 & 78.3 & - & -  & 256 & - & - & - & - & -\\
    \pz\pz ViT-VQGAN~\cite{yu2021vector} & 1.7B & 64M & 4.17 & 175.1 & - & -  & 1024 & - & - & - & - & -\\
    \pz\pz LlamaGen-3B~\cite{sun2024autoregressive} & 3.1B & 72M & 2.18 & 263.3 & 0.81 & 0.58  & 576 & - & - & - & - & -\\
    \pz\pz VAR ($d30$/$d36$-s)~\cite{tian2024visual} & 2B / 2.4B & 109M & 1.92 & 323.1 & 0.82 & 0.59  & 256 & 2.63 & 303.2 & - & - & 1024 \\
    \pz\pz MAR (H/L) ~\cite{li2024autoregressive} & 943M / 481M & 66M & 1.55 & 303.7 & 0.81 & 0.62  &256~\scriptsize{(d=16)} & 1.73 & 279.9 & - & - &1024~\scriptsize{(d=16)}\\
    \arrayrulecolor{gray}\cmidrule(lr){1-13}
    \multicolumn{6}{l}{\textit{latent diffusion}\vspace{0.02in}} \\
    \pz\pz LDM-4~\cite{rombach2022high} & 400M & 55M &3.60 & 247.7 & 0.87 & 0.48  & 4096~\scriptsize{(d=3)} & - & - & - & - & -\\
    \pz\pz U-ViT-H~\cite{bao2022all} & 501M & 84M & 2.29 & 263.9 & 0.82 & 0.57  &1024$^*$~\scriptsize{(d=4)} & 4.05 & 263.8 & 0.84 & 0.48 &4096$^*$~\scriptsize{(d=4)}\\
    \grayrow
    \pz\pz \textbf{DiT-XL/2}~\cite{peebles2023scalable} & 675M & 84M &2.27 & 278.2 & 0.83 & 0.57  & 1024$^*$~\scriptsize{(d=4)} &3.04 & 240.8 & 0.84 & 0.54  & 4096$^*$~\scriptsize{(d=4)}\\
    \pz\pz DiffiT~\cite{hatamizadeh2025diffit} & - & - & 1.73 & 276.5 & 0.80 & 0.62  & - & 2.67 & 252.1 & 0.83 & 0.55 & - \\
    \pz\pz MDTv2-XL/2~\cite{gao2023masked} & 676M & 84M & 1.58 & 314.7 & 0.79 & 0.65  & 1024$^*$~\scriptsize{(d=4)} & - & - & - & - & -\\
    \pz\pz {REPA + SiT-XL/2}~\cite{yu2024representation} & {675M} & {84M} & {1.80} & {284.0} & {0.81} & {0.61}  & {1024$^*$~\scriptsize{(d=4)}} & {-} & {-} & {-} & {-} & {-} \\
    \pz\pz EDM2-XXL~\cite{karras2024analyzing} & 1.5B & 84M & - & - & - & - & - & 1.81 & - & - & -  & 4096~\scriptsize{(d=4)}\\
    \arrayrulecolor{gray}\cmidrule(lr){1-13}
    \grayrow
    \multicolumn{13}{l}{\textit{Ours}} \\
    \grayrow
    \pz\pz \textbf{TexTok-32 + DiT-XL} & 675M & 176M & 2.75 & 294.6 & 0.83 & 0.56  & 32~\scriptsize{(d=8)} & 2.74 & 303.2 & 0.83 & 0.56 & 32~\scriptsize{(d=8)} \\
    \grayrow
    \pz\pz \textbf{TexTok-64 + DiT-XL} & 675M & 176M & 2.06 & 290.0 & 0.81 & 0.60  & 64~\scriptsize{(d=8)} & 1.99 & 301.9 & 0.82 & 0.6 & 64~\scriptsize{(d=8)} \\
    \grayrow
    \pz\pz \textbf{TexTok-128 + DiT-XL} & 675M & 176M & 1.66  & 294.4 & 0.80 & 0.61  & 128~\scriptsize{(d=8)} & 1.80  & 305.4 & 0.81 & 0.63 & 128~\scriptsize{(d=8)} \\
    \grayrow
    \pz\pz \textbf{TexTok-256 + DiT-XL} & 675M & 176M & \textbf{1.46} & 303.1 & 0.79 & 0.64  & 256~\scriptsize{(d=8)} & \textbf{1.62} & 313.8 & 0.80 & 0.64 & 256~\scriptsize{(d=8)} \\
    \bottomrule
    \end{tabular}
    } %
    \end{center}
    \vspace{-4mm}
    \caption{\footnotesize \textbf{System-level comparison of class-conditional image generation}  on ImageNet 256$\times$256 and 512$\times$512.
    TexTok-256 + DiT-XL achieves \textbf{\textit{state-of-the-art}} performance on both image resolutions. All entries \textit{use} classifier-free guidance if applicable. 
    Note that our method is orthogonal to both latent generation models and classifier-free guidance techniques, more advanced latent generators~\cite{ma2024sit} and guidance mechanisms~\cite{kynkaanniemi2024applying, karras2024guiding} can also be applied to TexTok to further improve our performance. 
    ``\#Params (G)": the number of generator's parameters. ``\#Params (T)": the number of tokenizer's parameters. ``\#tokens": the number of latent image tokens used during generation. ``/" in the first three columns indicates different configurations used at different image resolutions respectively. 
    $^*$ denotes the number of tokens before patchification.}
    \label{tab:imagenet-system}
\end{table*}

%% file: figure_src/inference_time_256_512.tex
\begin{figure*}[!h]
\begin{minipage}{0.66\textwidth}
    \centering
    \subfloat[ImageNet 256$\times$256]{%
        \includegraphics[width=0.49\linewidth]{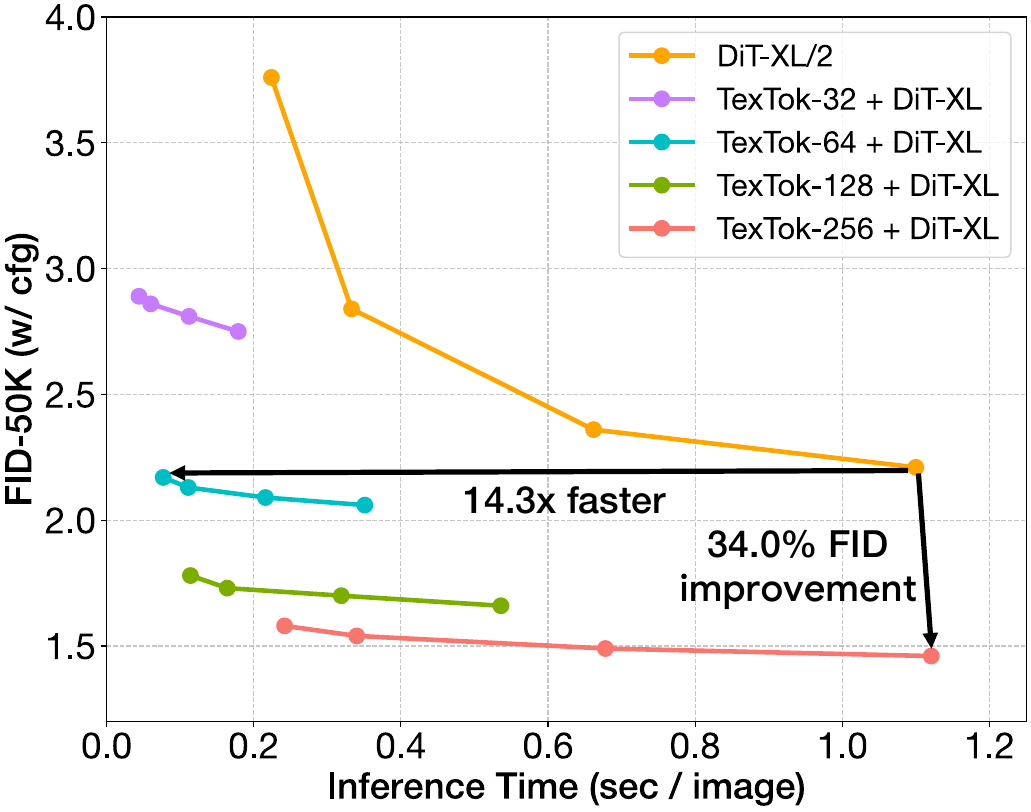}%
        \label{fig:inference_time_256}%
    }
    \hfill
    \subfloat[ImageNet 512$\times$512]{%
        \includegraphics[width=0.49\linewidth]{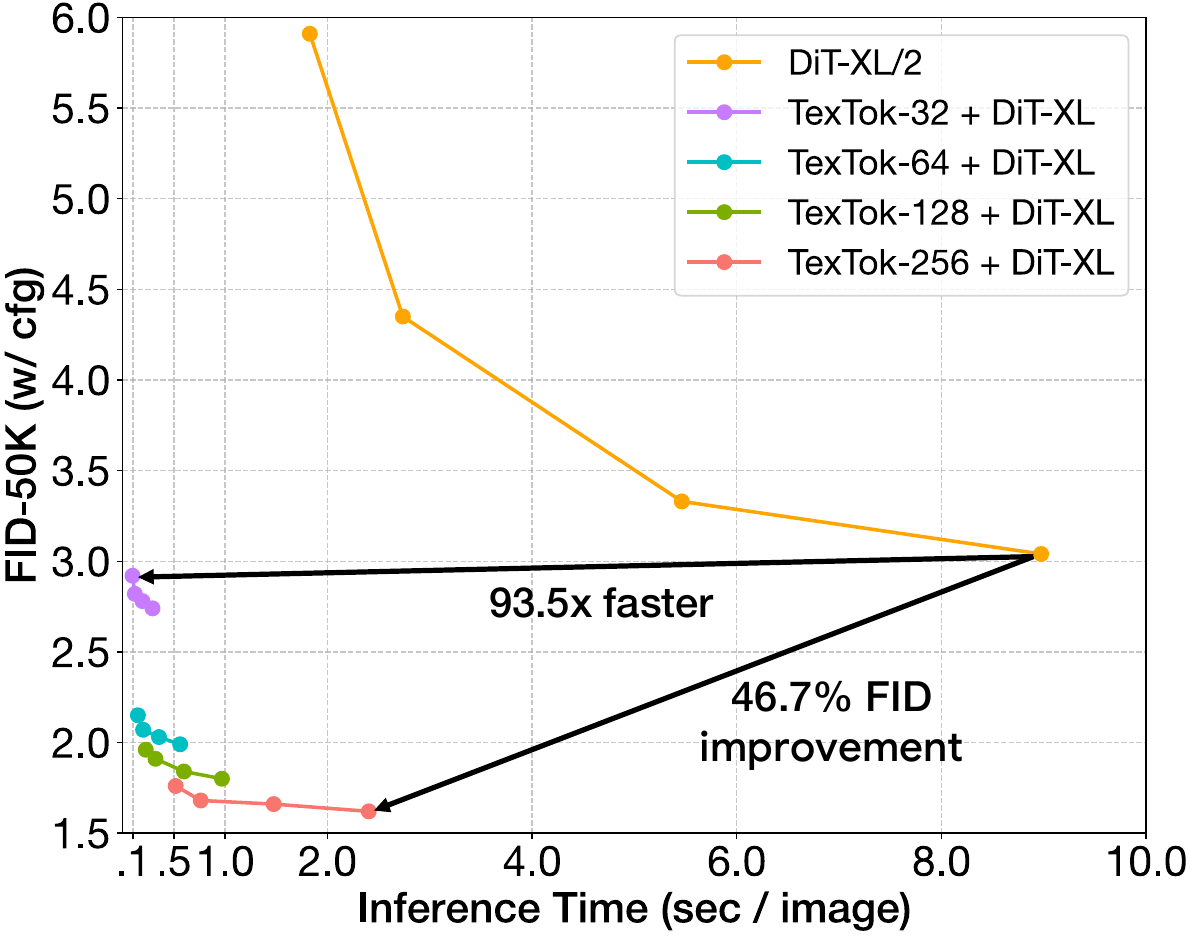}%
        \label{fig:inference_time_512}%
    }
    \caption{\footnotesize \textbf{Speed/performance tradeoff} of TexTok + DiT-XL compared to the original DiT-XL/2 on ImageNet 256$\times$256 and 512$\times$512. TexTok achieves the same generation performance 14.3$\times$/93.5$\times$ faster, or gains 34.0\%/46.7\% FID improvements using similar inference time. As image resolution scales up, this improvement is more pronounced.
    Each curve is obtained by using different sampling steps (50, 75, 150, 250). The inference time includes latent token generation, T5 text embedding extraction (for TexTok), and detokenization, measured on a single TPUv5e chip with a batch size of 32. }
    \label{fig:inference_time}
\end{minipage}%
\hfill
\begin{minipage}{0.32\textwidth}
\begin{table}[H]
    \begin{center}
    \footnotesize
    \begin{tabular}{lcc}
    \midrule
    Tokenizer  & FID$\downarrow$   & CLIP Score$\uparrow$ \\
    \midrule
     Baseline-32  & 5.09 & 28.08 \\
    \textbf{TexTok-32} & \textbf{4.36} & \textbf{28.73} \\
    \midrule
     Baseline-64  & 3.74 & 28.49 \\
    \textbf{TexTok-64} & \textbf{3.34} & \textbf{28.92} \\
    \midrule
    Baseline-128  & 3.01 & 28.95 \\
    \textbf{TexTok-128} & \textbf{2.82} & \textbf{29.23} \\
    \bottomrule
    \end{tabular}
    \end{center}
    \vspace{-3mm}
    \caption{\footnotesize \textbf{Text-to-image generation performance comparison of TexTok with \wotext{}} using DiT-XL-T2I on ImageNet 256$\times$256.
    TexTok achieves better FID and CLIP scores on all 32/64/128-token settings, indicating that TexTok produces image tokens that improve text-to-image generation results using the exactly \textit{same} image generation setups. Classifier-free guidance is applied.}  %
    \label{tab:t2i}
\end{table}
\end{minipage}
\end{figure*}

%% file: figure_src/t2i_samples.tex
\begin{figure*}[ht]\centering
\includegraphics[width=0.98\linewidth]{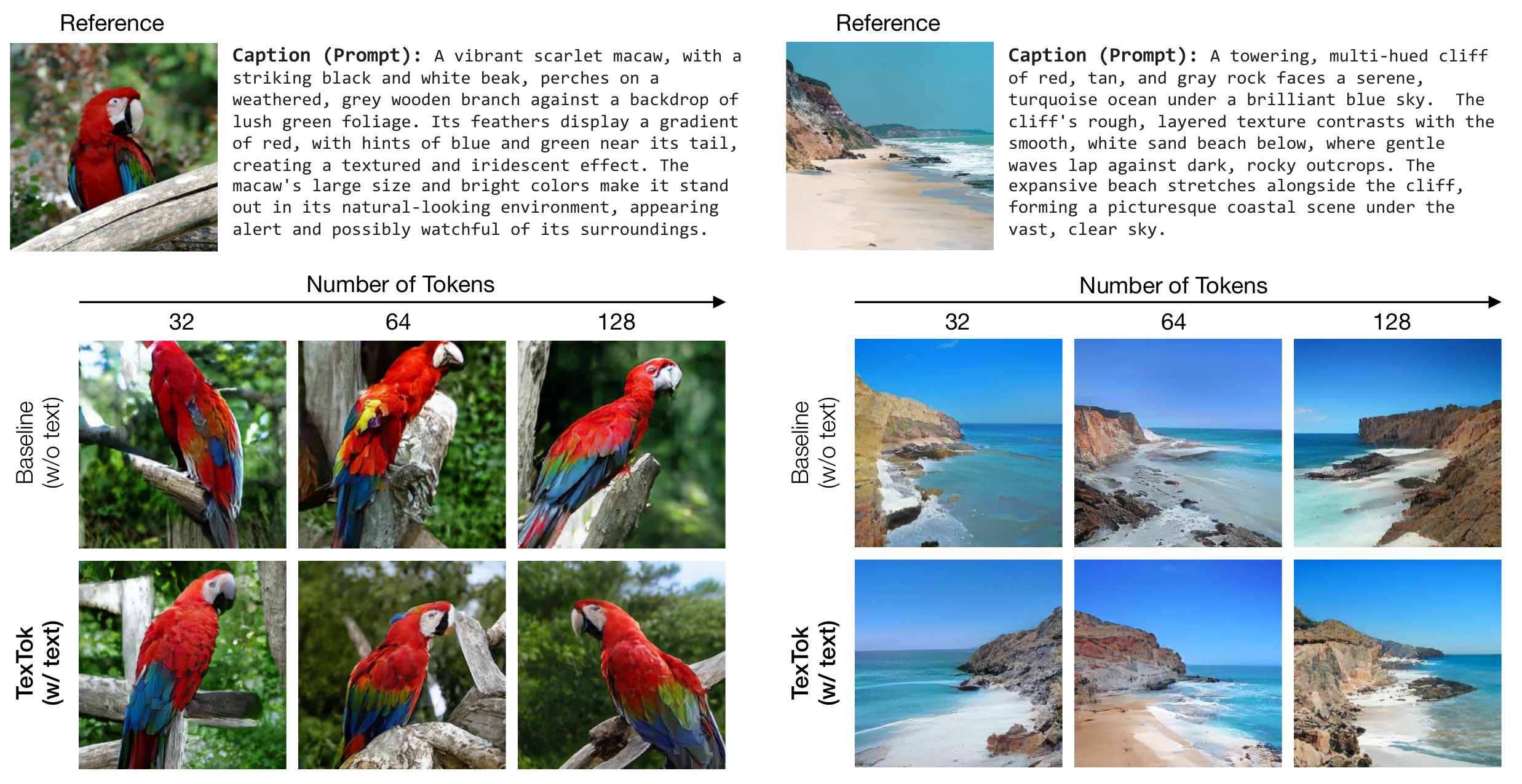}
\vspace{-4mm}
\caption{\footnotesize \textbf{Qualitative text-to-image generation results of TexTok compared with \wotext{}} on ImageNet 256$\times$256.
TexTok generates higher-quality images that better follow the prompts compared to \wotext{}. It even captures some fine-grained visual details presented in the reference images. The first row shows reference images from the ImageNet validation set along with their captions. Both TexTok and \wotext{} use the same generation settings and are conditioned on the same captions.
}
\label{fig:t2i_samples}
\vspace{-4mm}
\end{figure*}

%% file: figure_src/example_512.tex
\newcommand{\hhs}{\hspace{-0.001em}}
\newcommand{\vvs}{\vspace{-.1em}}
\begin{figure}[t]
\centering
\includegraphics[width=0.33\linewidth]{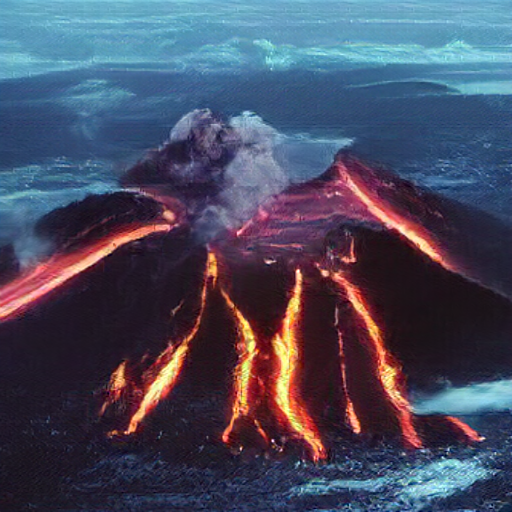}\hhs
\includegraphics[width=0.33\linewidth]{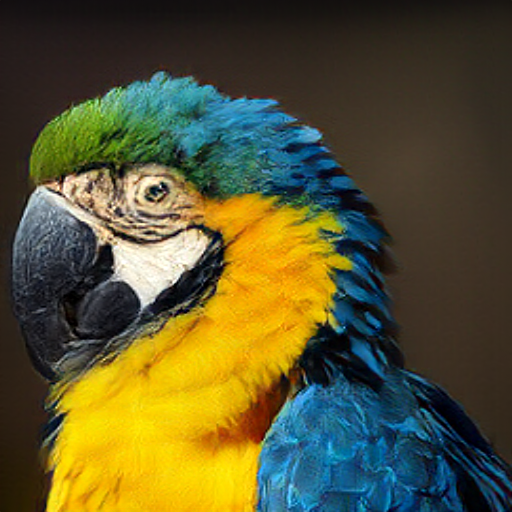}\hhs
\includegraphics[width=0.33\linewidth]{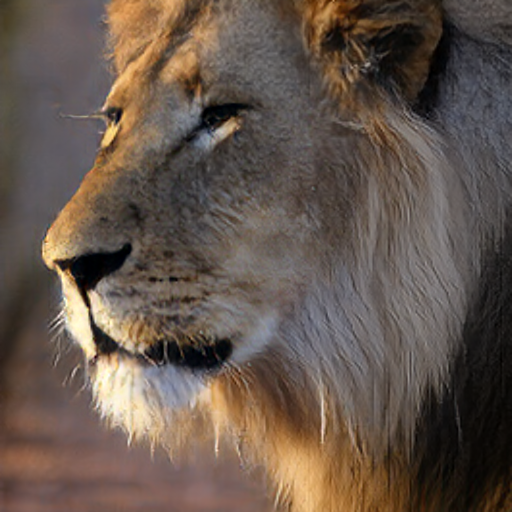}\vvs
\\
\includegraphics[width=0.165\linewidth]{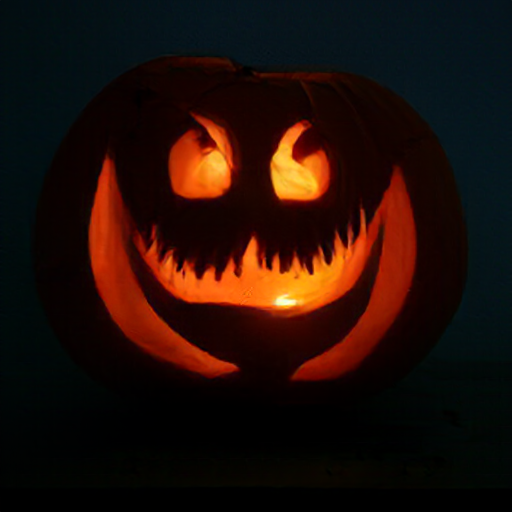}\hhs
\includegraphics[width=0.165\linewidth]{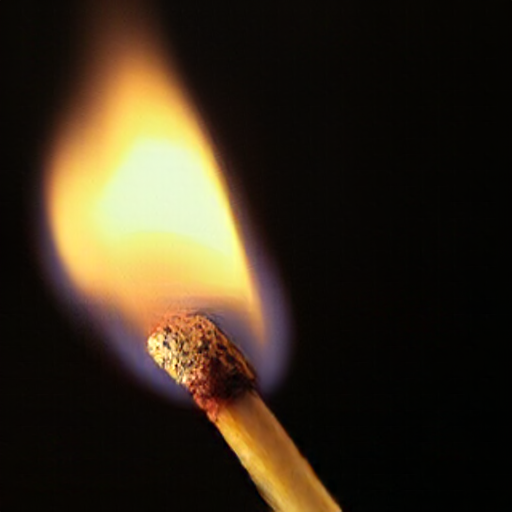}\hhs
\includegraphics[width=0.165\linewidth]{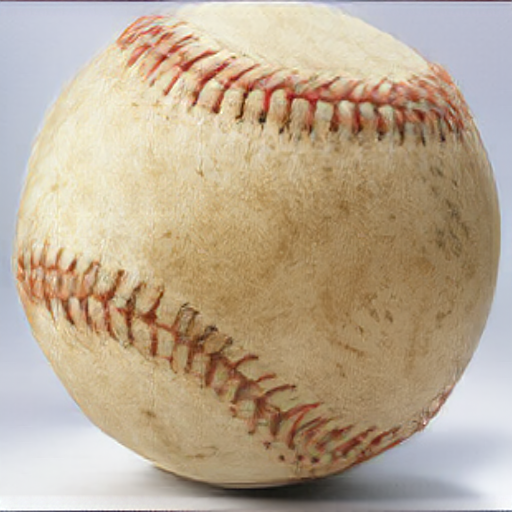}\hhs
\includegraphics[width=0.165\linewidth]{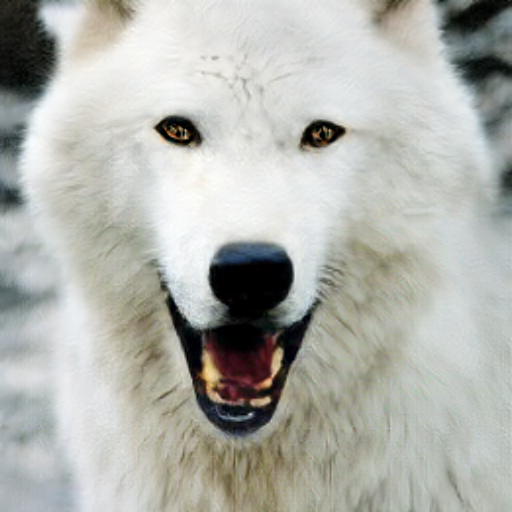}\hhs
\includegraphics[width=0.165\linewidth]{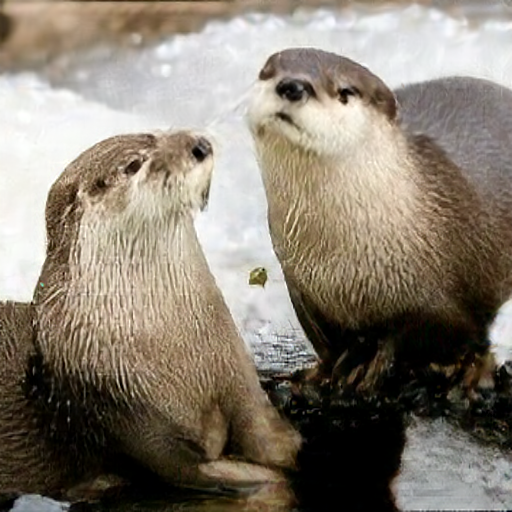}\hhs
\includegraphics[width=0.165\linewidth]{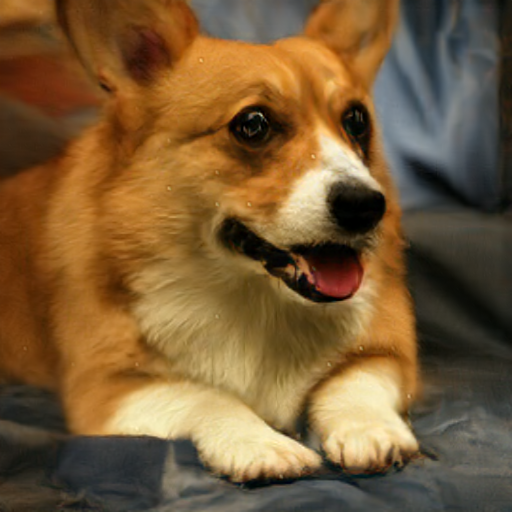}\vvs
\\
\includegraphics[width=0.165\linewidth]{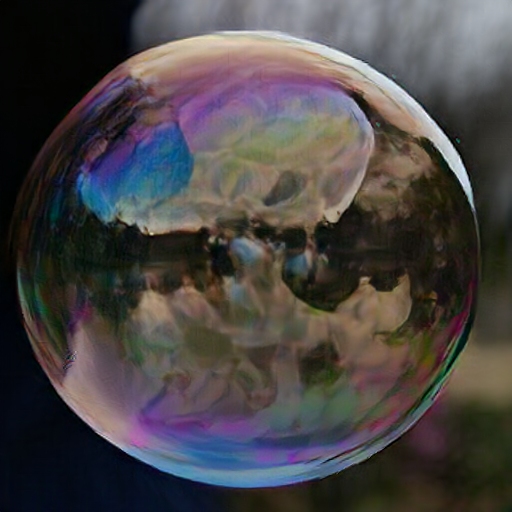}\hhs
\includegraphics[width=0.165\linewidth]{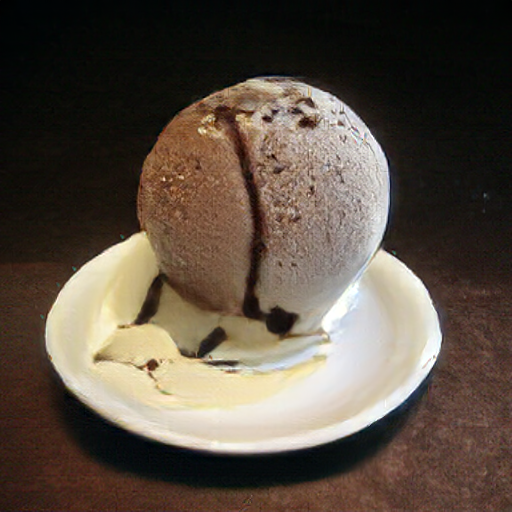}\hhs
\includegraphics[width=0.165\linewidth]{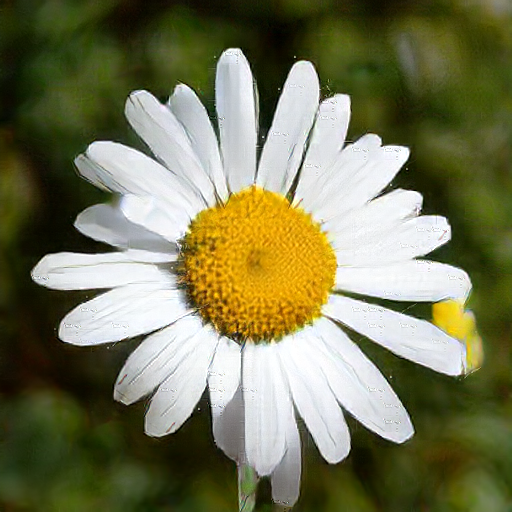}\hhs
\includegraphics[width=0.165\linewidth]{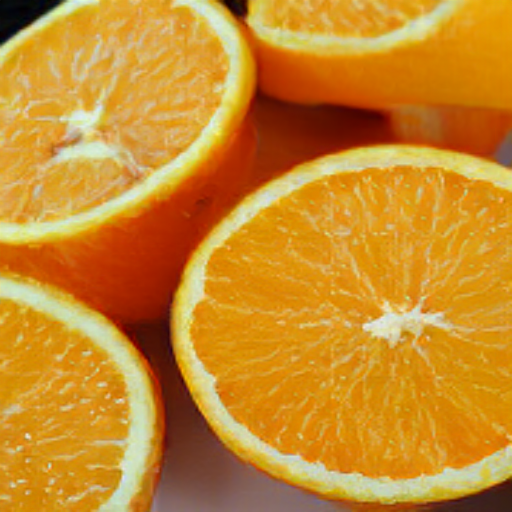}\hhs
\includegraphics[width=0.165\linewidth]{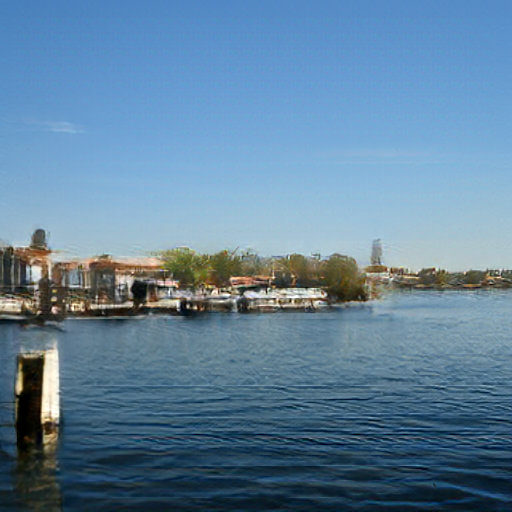}\hhs
\includegraphics[width=0.165\linewidth]{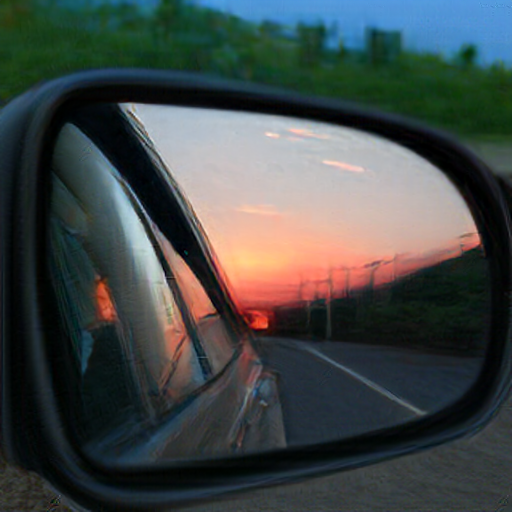}\vvs
\vspace{-1.5mm}
\caption{\footnotesize \textbf{Qualitative class-conditional image generation results} on ImageNet 512$\times$512.
TexTok generates semantically meaningful images with delicate fine-grained details.
Results are generated with our class-conditional TexTok-256 + DiT-XL model.}
\label{fig:teaser}
\vspace{-4mm}
\end{figure}

%% file: tables/tab-ablation.tex
\begin{table*}[t]
\vspace{-6mm}
\centering
\subfloat[
\scriptsize \textbf{Text conditioning}. More descriptive text captions yield better results.
\label{tab:cond_types}
]{
\centering
\begin{minipage}{0.29\linewidth}{\begin{center}
\scriptsize
\begin{tabular}{x{76}x{24}x{24}}
Text conditioning & rFID$\downarrow$ & PSNR$\uparrow$ \\
\midrule[1pt]
none & 1.49 & 20.51 \\
class category & 1.14 & 21.56 \\
class text & 1.15 & 21.58 \\
25-word caption & 1.08 & 21.63 \\
\grayrow
75-word caption & \textbf{1.04} & \textbf{22.05} \\
\end{tabular}
\end{center}}
\end{minipage}
}
\hspace{2em}
\subfloat[
\scriptsize
\textbf{T5 text encoder size}. Larger text encoder model size is better. 
\label{tab:t5_size}
]{
\begin{minipage}{0.29\linewidth}{\begin{center}
\scriptsize
\begin{tabular}{x{56}x{24}x{24}}
T5 model size & rFID$\downarrow$ & PSNR$\uparrow$ \\
\midrule[1pt]
Small & 1.06 & 22.01 \\
\grayrow
XL & 1.04 & 22.05 \\
XXL & \textbf{0.99} & \textbf{22.28} \\
\multicolumn{3}{c}{~}\\
\multicolumn{3}{c}{~}\\
\multicolumn{3}{c}{~}\\
\end{tabular}
\end{center}}\vspace{-6.3mm}\end{minipage}
}
\hspace{2em}
\subfloat[
\scriptsize
\textbf{Conditioning architecture}. In-context conditioning in the self-attention layers is better than adding a cross-attention layer in each ViT block.
\label{tab:cond_design}
]{
\begin{minipage}{0.29\linewidth}\vspace{-4mm}{\begin{center}
\scriptsize
\begin{tabular}{x{76}x{24}x{24}z{24}}
Conditioning architecture & rFID$\downarrow$ & PSNR$\uparrow$ \\
\midrule[1pt]
none & 1.49 & 20.51 \\
cross-attention layer & 1.31 & 21.42 \\
\grayrow
in-context conditioning & \textbf{1.04} & \textbf{22.05} \\
\multicolumn{3}{c}{~}\\
\multicolumn{3}{c}{~}\\
\multicolumn{3}{c}{~}\\
\multicolumn{3}{c}{~}\\
\end{tabular}
\end{center}}\vspace{-9.0mm}\end{minipage}
}
\\
\vspace{1mm}
\subfloat[
\scriptsize
\textbf{Conditioning location}. Injecting text conditioning to both tokenizer and detokenizer obtains the best results.
\label{tab:cond_location}
]{
\begin{minipage}{0.36\linewidth}{\begin{center}
\scriptsize
\begin{tabular}{x{86}x{24}x{24}z{24}}
Conditioning location & rFID$\downarrow$ & PSNR$\uparrow$ \\
\midrule[1pt]
none & 1.49 & 20.51 \\
tokenizer only & 1.38 & 21.29 \\
\grayrow
tokenizer \& detokenizer & \textbf{1.04} & \textbf{22.05} \\
\multicolumn{3}{c}{~}\\
\end{tabular}
\end{center}}\vspace{-6mm}\end{minipage}
}
\hspace{1em}
\subfloat[
\scriptsize
\textbf{TexTok model size}. TexTok-Base has the best performance/efficiency tradeoff. \label{tab:tokenizer_size}
]{
\begin{minipage}{0.55\linewidth}{\begin{center}
\scriptsize
\begin{tabular}{x{56}x{24}x{48}x{24}x{36}x{24}x{24}}
Model size & Layers & Hidden size & Heads & \#Params & rFID$\downarrow$ & PSNR$\uparrow$ \\
\midrule[1pt]
TexTok-Small & 8+8& 512 & 8 & 54M & 1.35 & 21.43 \\
\grayrow
TexTok-Base & 12+12& 768 & 12 & 176M & 1.04 & 22.05 \\
TexTok-Large & 24+24& 1024 & 16 & 612M & \textbf{1.03} & \textbf{22.09} \\
\multicolumn{3}{c}{~}\\
\end{tabular}
\end{center}}\vspace{-6mm}\end{minipage}
}
\centering
\vspace{-2mm}
\caption{\footnotesize \textbf{Ablation studies}. 
We ablate key design choices affecting TexTok's reconstruction performance on ImageNet 256$\times$256. \colorbox{baselinecolor}{Default} setting: TexTok-Base-128, 75-word captions, T5-XL text encoder, with in-context conditioning applied to both tokenizer and detokenizer.}
\label{tab:ablations} \vspace{-3mm}
\end{table*}

%% file: tables/tab-inference-time.tex
\begin{table}[t]
    \centering
    \scalebox{0.67}{
    \begin{tabular}{l|cc|cc}
    \toprule
    &\multicolumn{2}{c|}{\textbf{ImageNet 256$\times$256}} & \multicolumn{2}{c}{\textbf{ImageNet 512$\times$512}} \\
    \midrule\midrule
    tokenizer & Tokenization & Generation &  Tokenization & Generation \\
    \midrule
    SD-VAE-f8~\cite{rombach2022high} & 0.047 & \underline{0.289} & 0.182 & \underline{1.078} \\
    \midrule
    Baseline-32 & 0.051 & 0.030 & 0.169 & 0.031 \\
    TexTok-32 & 0.054 & 0.031 & 0.172 & 0.033 \\
    \midrule
    Baseline-64 & 0.051 & 0.066 & 0.171 & 0.067 \\
    TexTok-64 & 0.054 & 0.067 & 0.174 & 0.072 \\
    \midrule
    Baseline-128 & 0.052 & 0.110 & 0.175 & 0.109 \\
    TexTok-128 & 0.054 & 0.111 & 0.178 & 0.113 \\
    \midrule
    Baseline-256 & 0.052 & 0.289 & 0.181 & 0.282 \\
    TexTok-256 & 0.055 & 0.292 & 0.183 & 0.295 \\
    \bottomrule
    \end{tabular}
    }
    \vspace{-1mm}
    \caption{\footnotesize \textbf{Tokenization \& generation inference time}. We evaluate the tokenization and generation inference time of different tokenizers with a DiT-L generator. The tokenzation inference time includes T5 text embedding extraction (for TexTok), tokenization, and detokenization. The generation inference time includes latent token generation, T5 text embedding extraction (for TexTok), and detokenization. Both are measured on a single TPUv6e chip with a batch size of 32 (unit: second per image).}
    \label{tab:inference_time}
\vspace{-4mm}
\end{table}

%% file: sec/5_conclusion.tex
\vspace{-1mm}
\section{Conclusion}
We present \textit{Text-Conditioned Image Tokenization (TexTok)}, a new framework that leverages captions to guide the tokenizer in learning image semantics, allowing more learning capacity and token space to be allocated to capture visual details. TexTok significantly improves both reconstruction and generation performance, achieving state-of-the-art results in conditional and text-to-image generation on ImageNet with computational efficiency. By mitigating the trade-off between reconstruction quality and compression rate, TexTok enables more efficient image generation.

%% file: sec/6_acknowledgement.tex
\paragraph{Acknowledgements.} We thank David Minnen, Eirikur Agustsson, Qihang Yu, Peng Cao, José Lezama, Long Zhao, Haotian Tang, Tianhong Li, Xuhui Jia, Ruben Villegas and Xingyi Zhou for helpful discussions and valuable feedback. KZ and DK are partly funded by Wistron Corporation.
\par

%% file: sec/X_suppl.tex
\clearpage
\setcounter{page}{1}
\appendix

\section{Effectiveness of TexTok on Discrete Tokens}\label{app:discrete}
In the main paper, we demonstrate that TexTok works well with continuous tokens. In this section, we further validate the effectiveness of TexTok with Vector-Quantized (VQ) discrete tokens. We use a codebook size of 4096. As shown in Table~\ref{table:imagenet-256-discrete}, TexTok consistently delivers significant improvements over \wotext{} in reconstruction performance. As the number of tokens decreases, the performance gains are more pronounced. These results verify the effectiveness of TexTok on discrete tokens, highlighting its versatility as a universal tokenization framework. This inherent compatibility with both continuous and discrete tokens allows TexTok to seamlessly integrate with a wide range of generative models, including diffusion models, autoregressive models, and others.
\input{tables/tab-imagenet-256-discrete}
\input{figure_src/training_curve}
\input{figure_src/caption_examples}

\section{Additional Training Analysis}
In Figure~\ref{fig:training_curve}, we further provide the training reconstruction FID comparison (evaluated on 10K samples) of TexTok-32 \emph{v.s.} Baseline-32 (w/o text) on ImageNet 512$\times$512. From the figure, it is clear that TexTok training is more efficient and effective, achieving faster convergence and better reconstruction quality.

\section{Additional Implementation Details}\label{app:implementation}
We provide detailed default training hyperparameters for TexTok-$N$ as listed below:
\begin{itemize}[nosep, leftmargin=*]
    \item ViT encoder/decoder hidden size: $768$.
    \item ViT encoder/decoder number of layers: $12$.
    \item ViT encoder/decoder number of heads: $12$.
    \item ViT encoder/decoder MLP dimensions: $3072$.
    \item ViT patch size: $8$ for $256\times 256$ image resolution and $16$ for $512\times 512$.
    \item Discriminator base channels: $128$.
    \item Discriminator channel multipliers: $1, 2, 4, 4, 4, 4$ for $256\times 256$ image resolution, and $0.5, 1, 2, 4, 4, 4, 4$ for $512\times 512$.
    \item Discriminator starting iterations: $80,000$.
    \item Latent shape: $N\times 8$.
    \item Reconstruction loss weight: $1.0$.
    \item Generator loss type: Non-saturating.
    \item Generator adversarial loss weight: $0.1$.
    \item Discriminator gradient penalty: r1 with cost $10$.
    \item Perceptual loss weight: $0.1$.
    \item LeCAM weight: $0.0001$.
    \item Peak learning rate: $10^{-4}$.
    \item Learning rate schedule: linear warm up and cosine decay.
    \item Optimizer: Adam with $\beta_1=0$ and $\beta_2=0.99$.
    \item EMA model decay rate: $0.999$.
    \item Training epochs: $270$.
    \item Batch size: $256$.
\end{itemize}

We provide detailed default training and evaluation hyperparameters for DiT as listed below:
\begin{itemize}[nosep, leftmargin=*]
\item Patch size: $1$.
\item Peak learning rate: $5\times 10^{-4}$.
\item Learning rate schedule: linear warm up and cosine decay.
\item Optimizer: AdamW with $\beta_1=0.9$, $\beta_2=0.99$, and $0.01$ weight decay.
\item Diffusion steps: $1000$.
\item Noise schedule: Linear.
\item Diffusion $\beta_0 = 0.0001$, $\beta_{1000} = 0.02$.
\item Training objective: v-prediction.
\item Sampler: DDIM.
\item Sampling steps: $250$.
\item Training epochs: $350$.
\item Batch size: $1024$.
\end{itemize}

For classifier-free guidance, we adopt the guidance schedule from~\cite{gao2023masked} following prior arts~\cite{yu2024language, gupta2023photorealistic}. For the model configurations and other implementation details, please refer to the original DiT paper~\cite{peebles2023scalable}.

\section{ImageNet Captioning Details}\label{app:captioning}
We provide the prompt fed into Gemini that we used to generate the captions for images in ImageNet training and validation sets:

\renewcommand{\ttdefault}{lmtt}
\noindent \texttt{\small Describe an image of a \{class\_name\} from the ImageNet dataset in a single continuous detailed paragraph, using no more than \{word\_count\} words. Include descriptions of the appearance, colors, textures, size, environment, and any notable features or actions. Make sure to provide a vivid and engaging description that captures the essence of the \{class\_name\}. Output only the description without any additional words or commentary and the description must not exceed \{word\_count\} words. <image\_bytes>}

Figure~\ref{fig:caption_example} shows several captioning examples, including both 25-word and 75-word captions. 

\section{Additional Qualitative Results}\label{app:qualitative}
We present additional qualitative class-conditional image generation results on ImageNet 256$\times$256 in Figure~\ref{fig:class-conditional-256}. We observe that TexTok generates high-quality, semantically meaningful images with intricate fine-grained details.

We also present additional text-to-image generation results on ImageNet 256$\times$256 in Figure~\ref{fig:t2i_pair_samples}. Note that TexTok generates photo-realistic images that accurately align with the given prompts and even share many visual details with the reference images that the model has never seen, demonstrating both high fidelity and the capability to follow the text prompts.

\section{Additional Discussion on Related Work}
There have been some recent efforts in image tokenization methods~\cite{yu2024spae, liu2024language, zhu2024beyond, liang2024lg} that aim to align image tokens with textual semantics. Approaches like LQAE~\cite{liu2024language}, SPAE~\cite{yu2024spae}, and V2L~\cite{zhu2024beyond} map images into tokens derived from the codebooks of large language models (LLMs), while LG-VQ~\cite{liang2024lg} focuses on aligning the decoder features of image tokens with text representations. These methods are designed primarily for bridging visual and textual modalities to improve multi-modal understanding tasks. However, by aligning image tokens directly to textual semantic spaces, they often suffer from limited image reconstruction quality due to the inherent divergence between vision and language representations. Consequently, these approaches fail to achieve reasonable performance, or even do not report results, on standard image generation benchmarks, such as ImageNet class-conditional generation.

In contrast, our work introduces a novel image tokenization framework specifically designed for generative tasks. We are the first work that proposes conditioning the tokenization process directly on the text captions of images, a strategy typically reserved for the generation phase. Rather than enforcing a strict alignment between image tokens and text captions, our method leverages descriptive captions to provide a compact semantic representation. This simplifies semantic learning, allowing more learning capacity and token space to be allocated to capture fine-grained visual details. This complementary approach significantly improves the reconstruction quality and compression rate of tokenization. Furthermore, we demonstrate that our approach achieves state-of-the-art performance on standard ImageNet conditional generation benchmarks on both 256$\times$256 and 512$\times$512 image resolutions, establishing our method as a distinct advancement over existing works.

\input{figure_src/example_256}
\input{figure_src/t2i_pair_examples}

%% file: tables/tab-imagenet-256-discrete.tex
\begin{table}[H]
    \centering
    \begin{small}
    \scalebox{0.89}{
    \begin{tabular}{lcccccc}
    \toprule
    tokenizer & \# tokens & rFID $\downarrow$   & rIS$\uparrow$  & PSNR$\uparrow$     & SSIM $\uparrow$ & LPIPS$\downarrow$ \\
    \midrule
    Baseline-32 & \multirow{2}{*}{32} & 7.71 & 84.0 & 15.52 & 0.3822 & 0.4524 \\
    \textbf{TexTok-32} & & \textbf{4.11} & \textbf{141.4} & \textbf{16.52} & \textbf{0.4040} & \textbf{0.3855} \\
    \midrule
    Baseline-64 & \multirow{2}{*}{64} & 4.34 & 110.1 & 17.11 & 0.4200 & 0.3470 \\
    \textbf{TexTok-64} & & \textbf{2.50} & \textbf{161.8} & \textbf{18.06} & \textbf{0.4462} & \textbf{0.2933} \\
    \midrule
    Baseline-128 & \multirow{2}{*}{128} & 2.34 & 139.8 & 18.91 & 0.4737 & 0.2476 \\
    \textbf{TexTok-128} & & \textbf{1.76} & \textbf{167.9} & \textbf{19.96} & \textbf{0.4926} & \textbf{0.2166} \\
    \midrule
    Baseline-256 & \multirow{2}{*}{256} & 1.45 & 159.4 & 20.67 & 0.5371 & 0.1848 \\
    \textbf{TexTok-256} & & \textbf{1.17} & \textbf{180.3} & \textbf{21.56} & \textbf{0.5526} & \textbf{0.1594} \\
    \bottomrule
    \end{tabular}
    } %
    \end{small}
    \vspace{-2mm}
    \caption{\textbf{Reconstruction performance comparison of TexTok with Baseline (w/o text) using \textit{discrete} tokens} on ImageNet 256$\times$256. TexTok works well with discrete tokens. It consistently outperforms Baseline (w/o text) by a large margin in image reconstruction quality, achieving more pronounced gains as the number of tokens decreases.}
    \vspace{-4mm}
    \label{table:imagenet-256-discrete}
\end{table}

%% file: figure_src/training_curve.tex
\begin{figure}[H]\centering
\includegraphics[width=0.85\linewidth]{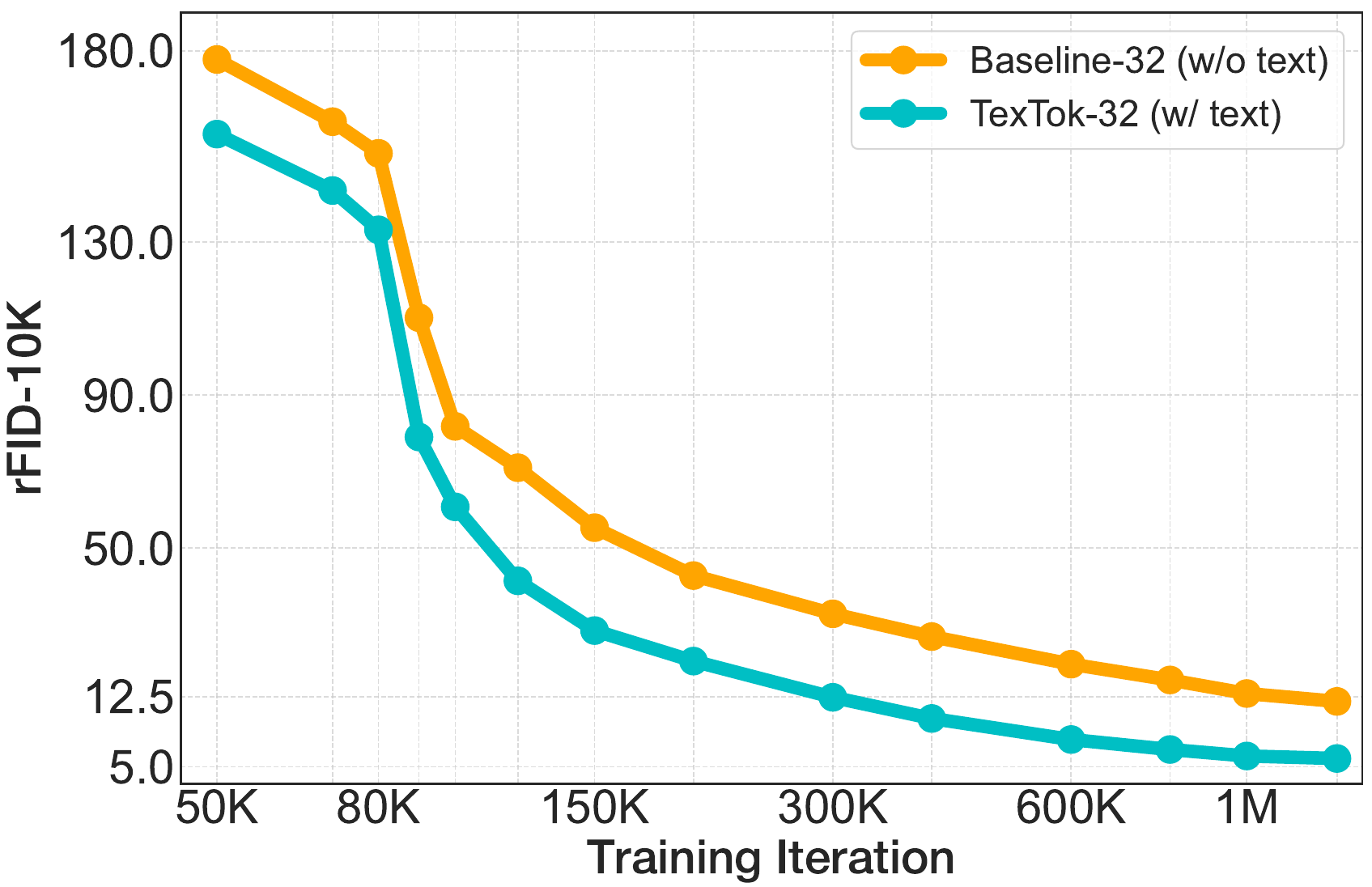}
\vspace{-2mm}
\caption{\textbf{Training reconstruction FID comparison of TexTok-32 \textit{v.s.} Baseline-32 (w/o text)} on ImageNet 512$\times$512. TexTok training is more efficient and effective, achieving faster convergence and better reconstruction quality.}
\vspace{-4mm}
\label{fig:training_curve}
\end{figure}

%% file: figure_src/caption_examples.tex
\begin{figure*}[t]\centering
\includegraphics[width=\textwidth]{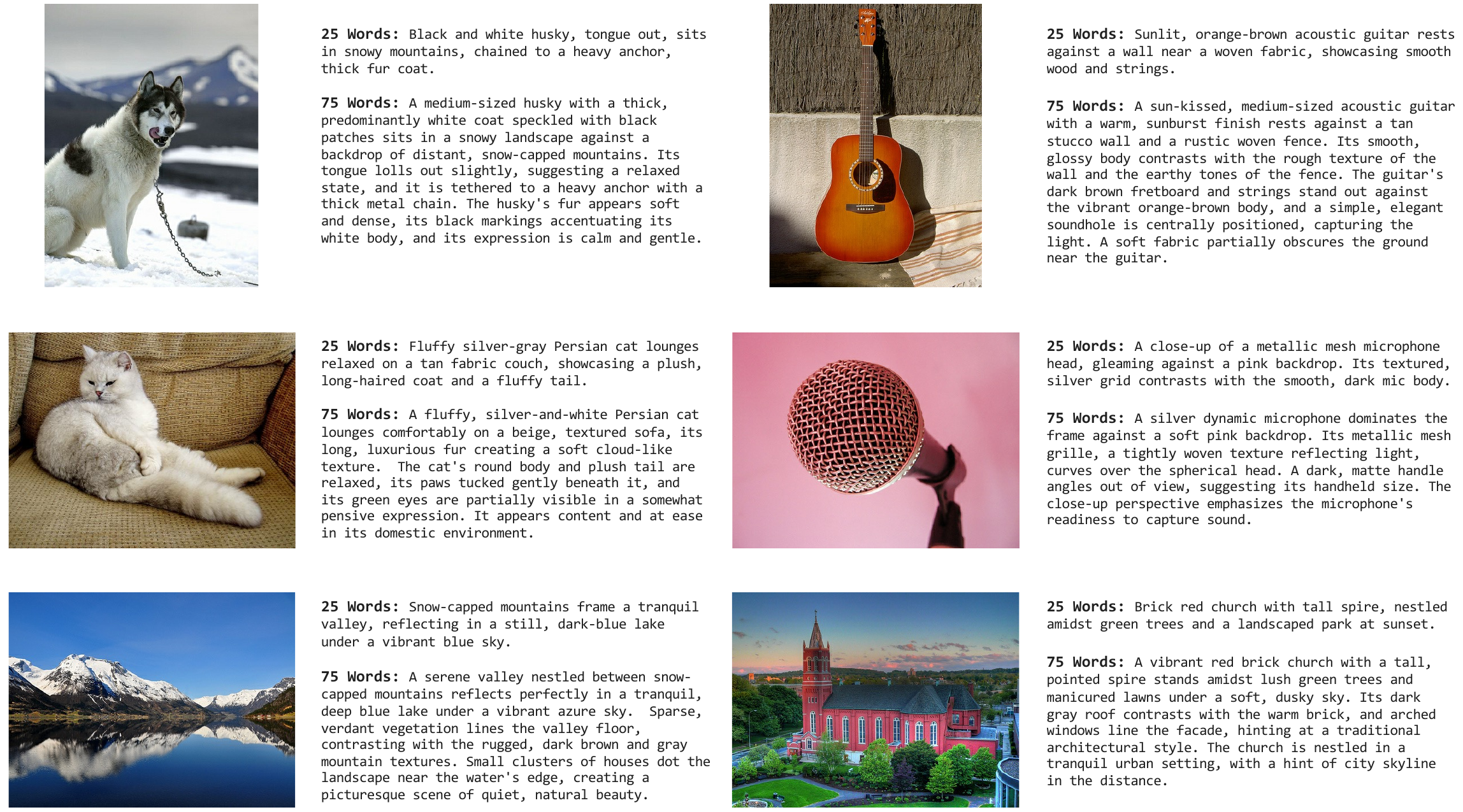}
\vspace{-6mm}
\caption{\textbf{ImageNet captioning examples}. Both 25-word and 75-word captions are displayed to the right of the corresponding images.}
\label{fig:caption_example}
\vspace{-3mm}
\end{figure*}

%% file: figure_src/example_256.tex
\begin{figure*}[t]
\vspace{-6mm}
\begin{center}
\centering
\includegraphics[width=0.33\linewidth]{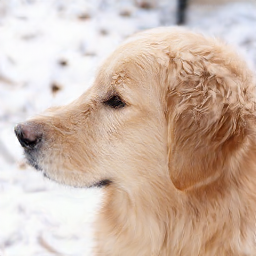}\hhs
\includegraphics[width=0.33\linewidth]{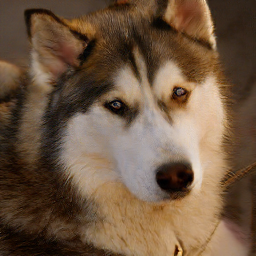}\hhs
\includegraphics[width=0.33\linewidth]{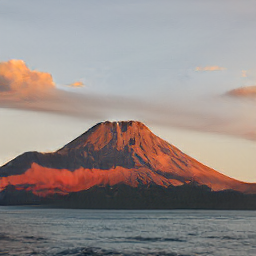}\vvs
\\
\includegraphics[width=0.165\linewidth]{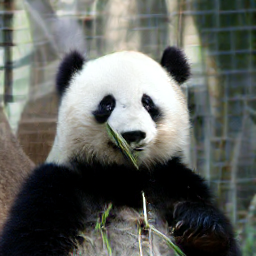}\hhs
\includegraphics[width=0.165\linewidth]{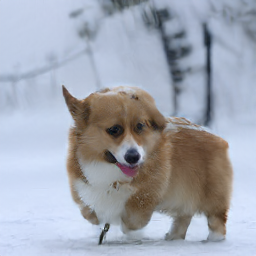}\hhs
\includegraphics[width=0.165\linewidth]{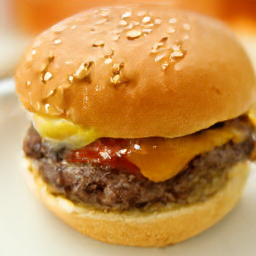}\hhs
\includegraphics[width=0.165\linewidth]{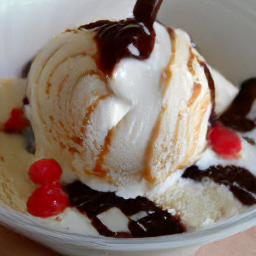}\hhs
\includegraphics[width=0.165\linewidth]{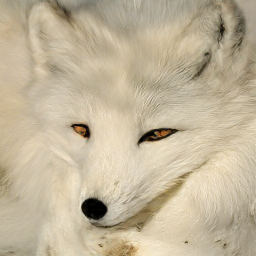}\hhs
\includegraphics[width=0.165\linewidth]{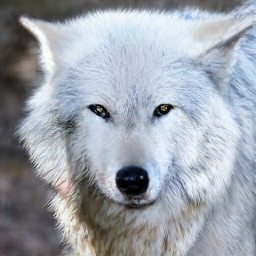}\vvs
\\
\includegraphics[width=0.165\linewidth]{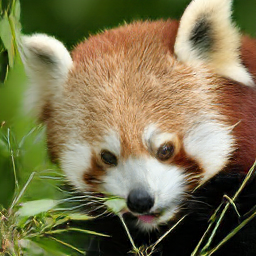}\hhs
\includegraphics[width=0.165\linewidth]{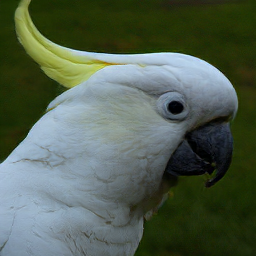}\hhs
\includegraphics[width=0.165\linewidth]{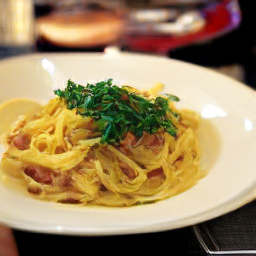}\hhs
\includegraphics[width=0.165\linewidth]{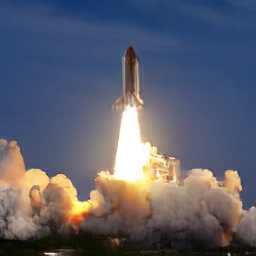}\hhs
\includegraphics[width=0.165\linewidth]{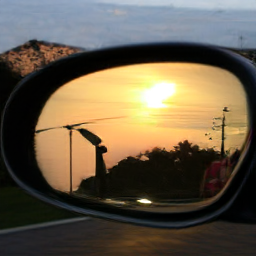}\hhs
\includegraphics[width=0.165\linewidth]{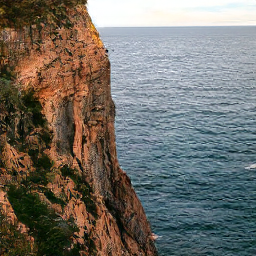}\vvs
\vspace{-1mm}
\caption{\textbf{Qualitative class-conditional image generation results} on ImageNet 256$\times$256.
TexTok generates high-quality, semantically meaningful images with intricate fine-grained details.
Results are generated with our class-conditional TexTok-256 + DiT-XL model.}
\label{fig:class-conditional-256}
\end{center}
\end{figure*}

%% file: figure_src/t2i_pair_examples.tex
\begin{figure*}[ht]\centering
\includegraphics[width=\textwidth]{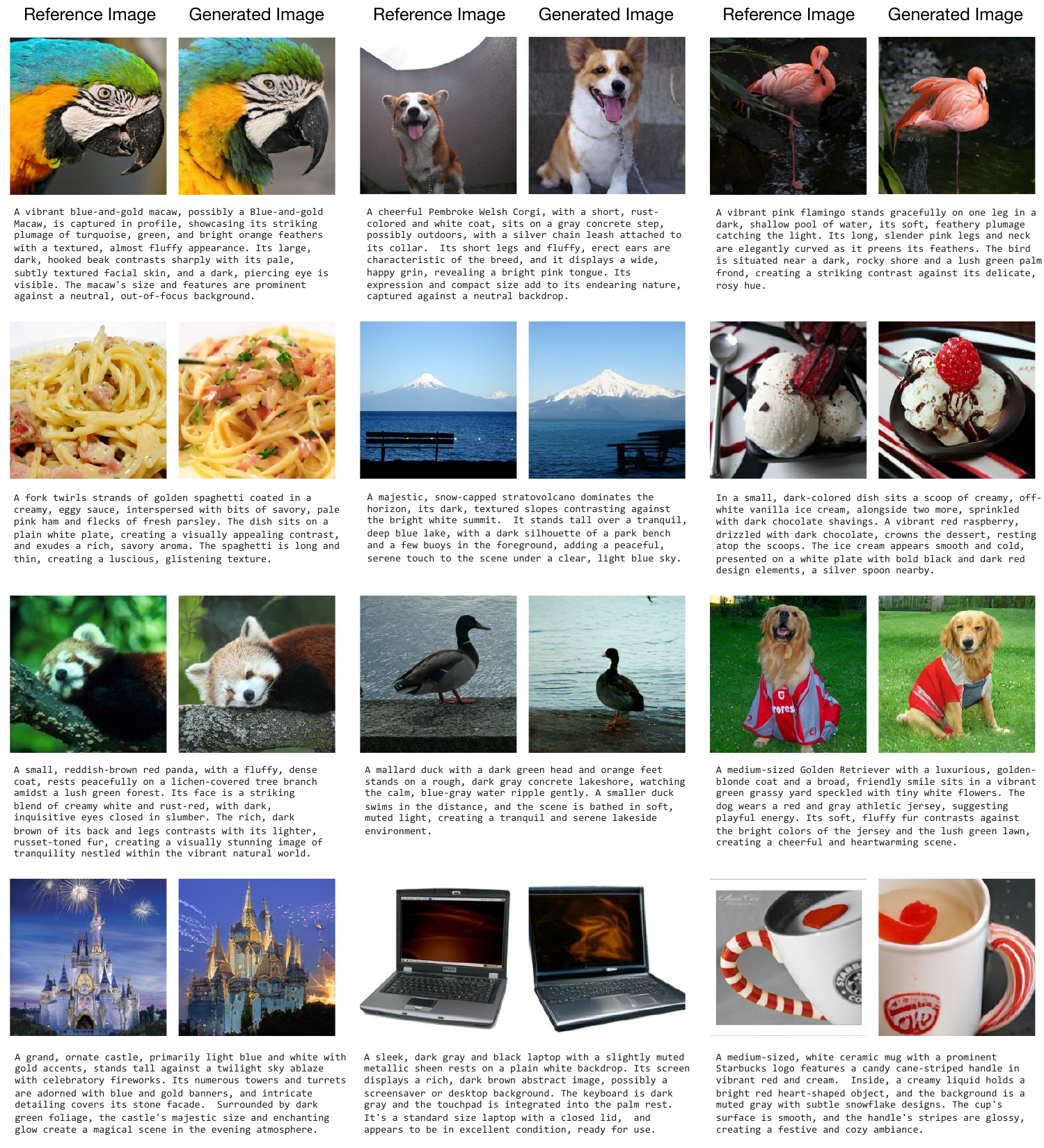}
\vspace{-6mm}
\caption{\textbf{Qualitative text-to-image generation results} on ImageNet 256$\times$256. For each sample, we present the reference image from the ImageNet validation set alongside the corresponding generated image, which is produced conditioned on the caption (displayed below) of the reference image. Results are generated with our text-to-image TexTok-128 + DiT-XL-T2I model. TexTok generates visually realistic images that accurately align with the given prompts, demonstrating both high fidelity and semantic relevance.
}
\label{fig:t2i_pair_samples}
\end{figure*}